\documentclass[journal]{IEEEtran}

\ifCLASSINFOpdf
\else
\fi

\usepackage{bm}
\usepackage{graphicx}
\usepackage{subfigure}
\usepackage{epstopdf}
\usepackage{amsmath}
\usepackage{amssymb}
\usepackage{cases}
\usepackage{algorithm}
\usepackage{algorithmic}
\usepackage{cite}
\usepackage{enumerate}
\usepackage{multirow}
\usepackage{makecell}
\usepackage{color}
\usepackage{romannum}
\usepackage[super]{nth}

\allowdisplaybreaks[1]
\usepackage{tabularx}
\usepackage{isomath}
\DeclareMathOperator{\tr}{tr}

\begin{document}
%
% paper title
% Titles are generally capitalized except for words such as a, an, and, as,
% at, but, by, for, in, nor, of, on, or, the, to and up, which are usually
% not capitalized unless they are the first or last word of the title.
% Linebreaks \\ can be used within to get better formatting as desired.
% Do not put math or special symbols in the title.
\title{Tensor Decomposition for Model Reduction in Neural Networks: A Review\thanks{This research was supported in part by the National Science Foundation under grant number CCF-1954749.}}
%
%
% author names and IEEE memberships
% note positions of commas and nonbreaking spaces ( ~ ) LaTeX will not break
% a structure at a ~ so this keeps an author's name from being broken across
% two lines.
% use \thanks{} to gain access to the first footnote area
% a separate \thanks must be used for each paragraph as LaTeX2e's \thanks
% was not built to handle multiple paragraphs
%

\author{Xingyi Liu, {\em Graduate Student Member, IEEE}; and Keshab K. Parhi, {\em Fellow, IEEE}\\
University of Minnesota\\
Department of Electrical and Computer Engineering\\
Minneapolis, MN, USA}

\maketitle

% As a general rule, do not put math, special symbols or citations
% in the abstract or keywords.
\begin{abstract}
Modern neural networks have revolutionized the fields of computer vision (CV) and Natural Language Processing (NLP). They are widely used for solving complex CV tasks and NLP tasks such as image classification, image generation, and machine translation. Most state-of-the-art neural networks are over-parameterized and require a high computational cost. One straightforward solution is to replace the layers of the networks with their low-rank tensor approximations using different tensor decomposition methods. This paper reviews six tensor decomposition methods and illustrates their ability to compress model parameters of convolutional neural networks (CNNs), recurrent neural networks (RNNs) and Transformers. The accuracy of some compressed models can be higher than the original versions. Evaluations indicate that tensor decompositions can achieve significant reductions in model size, run-time and energy consumption, and are well suited for implementing neural networks on edge devices.
\end{abstract}

% Note that keywords are not normally used for peerreview papers.
\begin{IEEEkeywords}
Tensor decomposition, convolution neural network acceleration, recurrent neural network acceleration, Transformer acceleration, Canonical Polyadic decomposition, Tucker decomposition, Tensor Train decomposition, Tensor Ring decomposition, Block-Term decomposition, Hierarchical Tucker decomposition, model compression.
\end{IEEEkeywords}

% For peer review papers, you can put extra information on the cover
% page as needed:
% \ifCLASSOPTIONpeerreview
% \begin{center} \bfseries EDICS Category: 3-BBND \end{center}
% \fi
%
% For peerreview papers, this IEEEtran command inserts a page break and
% creates the second title. It will be ignored for other modes.
\IEEEpeerreviewmaketitle

\section{Introduction}
\begin{figure}[htbp]
\centering
\resizebox{0.2\textwidth}{!}{%
\includegraphics{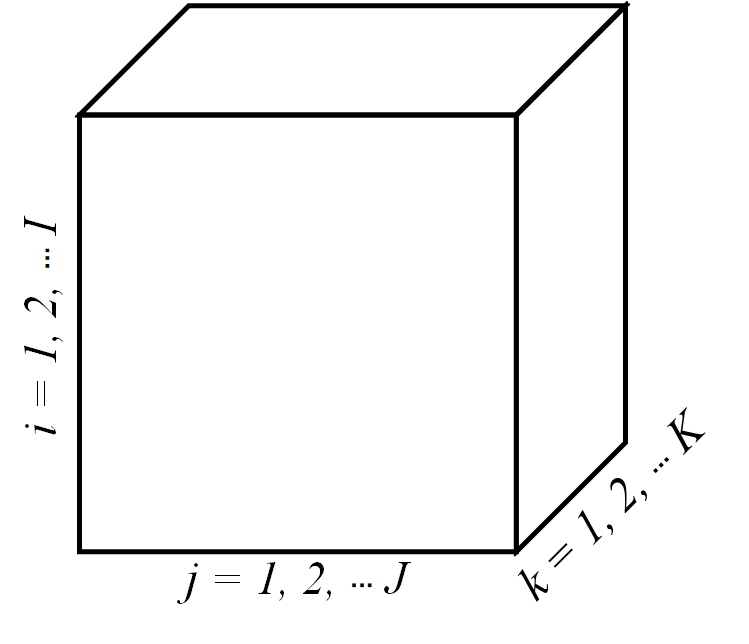}}
\caption{A third-order tensor: $\mathcal{X} \in \mathbb{R}^{I\times J\times K}$.}
\label{fig:tensor}
\end{figure}

\IEEEPARstart{T}{ensors} are multidimensional arrays indexed by three or more indices. An $N^{th}$-order tensor is the tensor product of $N$ vector spaces. Third-order tensors have three indices as shown in Fig.~\ref{fig:tensor}. In special cases, first-order tensors represent vectors, and second-order tensors represent matrices.

Convolutional neural networks (CNNs) have outperformed traditional techniques for image recognition tasks. In 2012, AlexNet~\cite{krizhevsky2012imagenet} achieved about $80 \%$ top-5 accuracy on the ImageNet dataset~\cite{deng2009imagenet}. Furthermore, subsequently VGG~\cite{simonyan2014very} and GoogleNet~\cite{szegedy2015going} achieved about $90 \%$ top-5 accuracy with the same dataset. On the ImageNet dataset, ResNet~\cite{he2016deep} with a depth of up to $152$ layers achieved $3.57 \%$ top-5 error. Executing CNNs for computer vision tasks on mobile devices is gaining more and more attention. Common methods to reduce the size of CNNs include: sparsification~\cite{figurnov2016perforatedcnns,han2015learning,molchanov2017variational,deng2018permdnn}, quantization~\cite{bulat2019matrix,krizhevsky2012imagenet}, structural pruning~\cite{gao2018dynamic,he2018soft,he2018amc,zhuang2018discrimination}, and low-rank approximation~\cite{astrid2017cp,denton2014exploiting,lebedev2014speeding,gusak2019automated,kim2015compression,novikov2015tensorizing,garipov2016ultimate,sidiropoulos2017tensor}.

The use of low-rank approximations is inspired by~\cite{denil2013predicting} which showed that the neural network parameters are highly redundant. The authors in this paper could predict more than $95 \%$ of the weights of the network which indicates that the parameters of the CNNs are highly over-parametrized. Various low-rank tensor/matrix decompositions can be directly applied to the weights of convolutional and fully connected layers. In this paper, we review Canonical Polyadic decomposition (CPD)~\cite{astrid2017cp,denton2014exploiting,lebedev2014speeding}, Tucker decomposition~\cite{gusak2019automated,kim2015compression} and Tensor Train decomposition~\cite{novikov2015tensorizing,garipov2016ultimate} approaches to reduce model parameters of CNNs. The decomposed layers are replaced by a sequence of new layers with significantly fewer parameters. 

Recurrent Neural Networks (RNNs) have shown promising success in sequence modeling tasks due to their ability in capturing temporal dependencies from input sequences~\cite{rumelhart1986learning,sutskever2014sequence}. Their advanced variant, the Long Short-Term Memory (LSTM), introduces a number of gates and passes information with element-wise operations to address the gradient vanishing issue in vanilla RNNs~\cite{hochreiter1997long}. These neural network architectures have shown promising performances in Natural Language Processing (NLP)~\cite{sutskever2011generating}, speech recognition~\cite{mikolov2011extensions} and computer vision (CV)~\cite{yu2016video}. The reader is referred to~\cite{parhi2020brain} for a review of various brain-inspired computing models.

Despite the success, RNNs and LSTMs suffer from a huge number of parameters which make the models notoriously difficult to train and susceptible to overfitting. In order to circumvent this problem, current approaches often involve exploring the low-rank structures of the weight matrices. Inspired by the implementation of tensor decomposition methods in CNNs~\cite{novikov2015tensorizing}, various tensor decomposition methods have been applied in RNNs and LSTMs, including Tensor Train Decomposition~\cite{yang2017tensor}, Tensor Ring Decomposition~\cite{pan2019compressing}, Block-Term Decomposition~\cite{ye2018learning} and Hierarchical Tucker Decomposition~\cite{yin2020compressing}. These tensor-decomposed models can maintain high performance with orders-of-magnitude fewer parameters compared to the large-size vanilla RNNs/LSTMs.

Transformer is a deep learning model that is based on the mechanism of self-attention by weighting the significance of each part of the input data differentially. It has led to breakthroughs in the fields of NLP and CV. Like RNNs, transformers are designed to process sequential input. However, transformers process the entire input all at once. For example, the transformers can process the whole natural language sentence at a time while the RNNs have to process word by word. The training parallelization allows transformers to be trained on larger datasets. This has led to the success of pre-trained systems such as BERT~\cite{devlin2018bert}, GPT~\cite{radford2018improving} and T5~\cite{raffel2020exploring}. However, the large model size of the Transformer based model may cause problems in training and inference under resource-limited environments. Tensorized embedding (TE) utilized the Tensor Train decomposition to compress the embedding layers of Transformers~\cite{hrinchuk2019tensorized}. A novel self-attention model with Block-Term Decomposition was proposed to compress the attention layers of Transformers~\cite{ma2019tensorized}. This method can not only largely compress the model size but also achieve performance improvement.

% This paper is organized as follows. Section~\ref{section:fc} briefly introduces matrix decomposition and tensor train decomposition for the fully connected layers of CNNs. Section~\ref{section:conv} reviews three tensor decomposition methods (Canonical Polyadic decomposition, Tucker decomposition and Tensor Train decomposition) that compress the convolutional layers of CNNs.
% Section~\ref{section:RNN} presents four tensor decomposition methods (Tensor Train decomposition, Tensor Ring decomposition, Block-Term decomposition and Hierarchical Tucker decomposition) that compress the feed-forward layers of RNNs. Implementations and results of applying these tensor decomposition methods to some classical CNN and RNN models are described in Section~\ref{section:model}. Finally, Section~\ref{section:conclusion} concludes the paper.

% Possible future directions of tensor decomposition are discussed in Section~\ref{section:future}.

\section{Tensor Decomposition Methods}
\label{section:td methods}
A tensor decomposition is any scheme for compressing a tensor into a sequence of other, often simpler tensors. In this section, we review some tensor decomposition methods that are commonly used to compress deep learning models.

\subsection{Truncated Singular Value Decomposition}
Given a matrix $\bm{W} \in \mathbb{R}^{M \times N}$, the singular value decomposition (SVD) of the matrix is defined as:

\begin{equation}
\centering
\bm{W}=\bm{U}\bm{S}\bm{V}^\intercal,\nonumber
\end{equation}
where $\bm{U} \in \mathbb{R}^{M \times M}$, $\bm{S} \in \mathbb{R}^{M \times N}$ and $\bm{V} \in \mathbb{R}^{N \times N}$. $\bm{S}$ is the diagonal matrix with all the singular values on the diagonal. $\bm{U}$ and $\bm{V}$ are the corresponding orthogonal matrices. If the singular values of $\bm{W}$ decay fast, then the weight matrix ($\bm{W}$) can be approximated by keeping only the $K$ largest entries of $\bm{S}$:

\begin{equation}
\centering
\widetilde{\bm{W}}=\widetilde{\bm{U}}\widetilde{\bm{S}}\widetilde{\bm{V}}^\intercal,\nonumber
\end{equation}
where $\widetilde{\bm{U}} \in \mathbb{R}^{M \times K}$, $\widetilde{\bm{S}} \in \mathbb{R}^{K \times K}$ and $\widetilde{\bm{V}} \in \mathbb{R}^{K \times N}$. Then for any $\bm{I} \in \mathbb{R}^{T \times M}$, the SVD approximation error satisfies:

\begin{equation}
\centering
||\bm{I}\widetilde{\bm{W}}-\bm{I}\bm{W}||_F \leq s_{K+1}||\bm{I}||_F,\nonumber
\end{equation}
where $||\cdot||_F$ denotes its Frobenius norm. Notice that the approximation error $||\bm{I}\widetilde{\bm{W}}-\bm{I}\bm{W}||_F$ is controlled by $s_{K+1}$, the $(K+1)^{th}$ largest singular value, or put another way, the decay along the diagonal of $\bm{S}$. Considering the computation cost, $\bm{I}\widetilde{\bm{W}}$ can be computed in $\mathcal{O}(TMK+TK^2+TKN)$ which is much smaller than $\mathcal{O}(TMN)$ for a sufficiently small $K$.

\subsection{Tensor Train Decomposition}
As defined in~\cite{oseledets2011tensor}, the Tensor Train (TT) Decomposition of a tensor $\mathcal{A} \in \mathbb{R}^{n_1\times n_2\times \dots \times n_d}$ can be replaced by a set of matrices $\bm{G}_k[j_k] \in \mathbb{R}^{r_{k-1}\times r_k}$ where $j_k = 1, 2, \dots, n_k$, $k = 1, 2, \dots, d$ and $r_0 = r_d = 1$. Then each of the tensor elements can be computed as:

\begin{equation}
\centering
\mathcal{A}(j_1, j_2, \dots, j_d) = \bm{G}_1[j_1]\bm{G}_2[j_2]\dots \bm{G}_d[j_d].\nonumber
\end{equation}

\begin{figure*}[htbp]
\centering
\resizebox{0.7\textwidth}{!}{%
\includegraphics{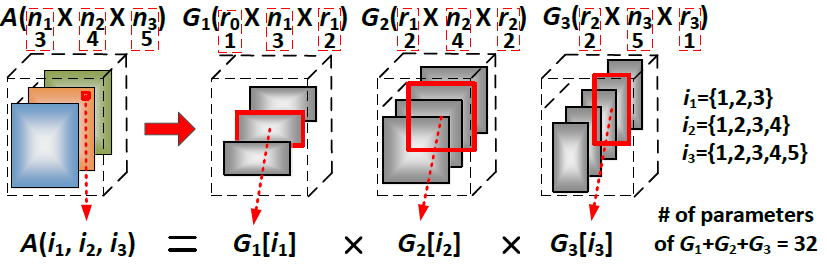}}
\caption{Tensor Train Decomposition of a third-order tensor~\cite{deng2019tie}. The original tensor has the dimensions $3 \times 4 \times 5$. After decomposition, the dimensions of the three TT-cores are $1 \times 3 \times 2$, $2 \times 4 \times 2$ and $2 \times 5 \times 1$, respectively. The number of parameters needed to represent the original tensor is reduced from $60$ to $32$.}
\label{fig:TT}
\end{figure*}

The sequence $\{r_k\}_{k=1}^d$ is called TT-rank of the TT-representation of $\mathcal{A}$. The collections of the matrices $\{\{\bm{G}_k[j_k]\}_{j_k=1}^{n_k}\}_{k=1}^d$ are defined as TT-cores~\cite{oseledets2011tensor}. Notice that in TT-format, only $\sum_{k=1}^d n_k r_{k-1} r_k$ parameters are required to represent a tensor $\mathcal{A} \in \mathbb{R}^{n_1 \times n_2 \times \dots \times n_d}$ which originally has $\prod_{k=1}^d n_k$ elements. The trade-off between the model compression ratio and the reconstruction accuracy is controlled by the TT-ranks ($\{r_k\}_{k=1}^d$). The smaller the TT-ranks, the higher the model compression ratio TT-format can achieve. Another advantage of the TT-decomposition is that basic linear algebra operations can be applied to the TT-format tensors efficiently~\cite{oseledets2011tensor}. Fig.~\ref{fig:TT} from~\cite{deng2019tie} shows that a third tensor $\mathcal{A} \in \mathbb{R}^{3\times 4\times 5}$ can be represented by three TT-cores ($\{\bm{G}_d\}_{d=1}^3$) with $32$ parameters in TT-format. Thus, the number of parameters needed to represent the original tensor is reduced from $60$ to $32$.

\begin{figure*}[htbp]
\centering
\resizebox{0.7\textwidth}{!}{%
\includegraphics{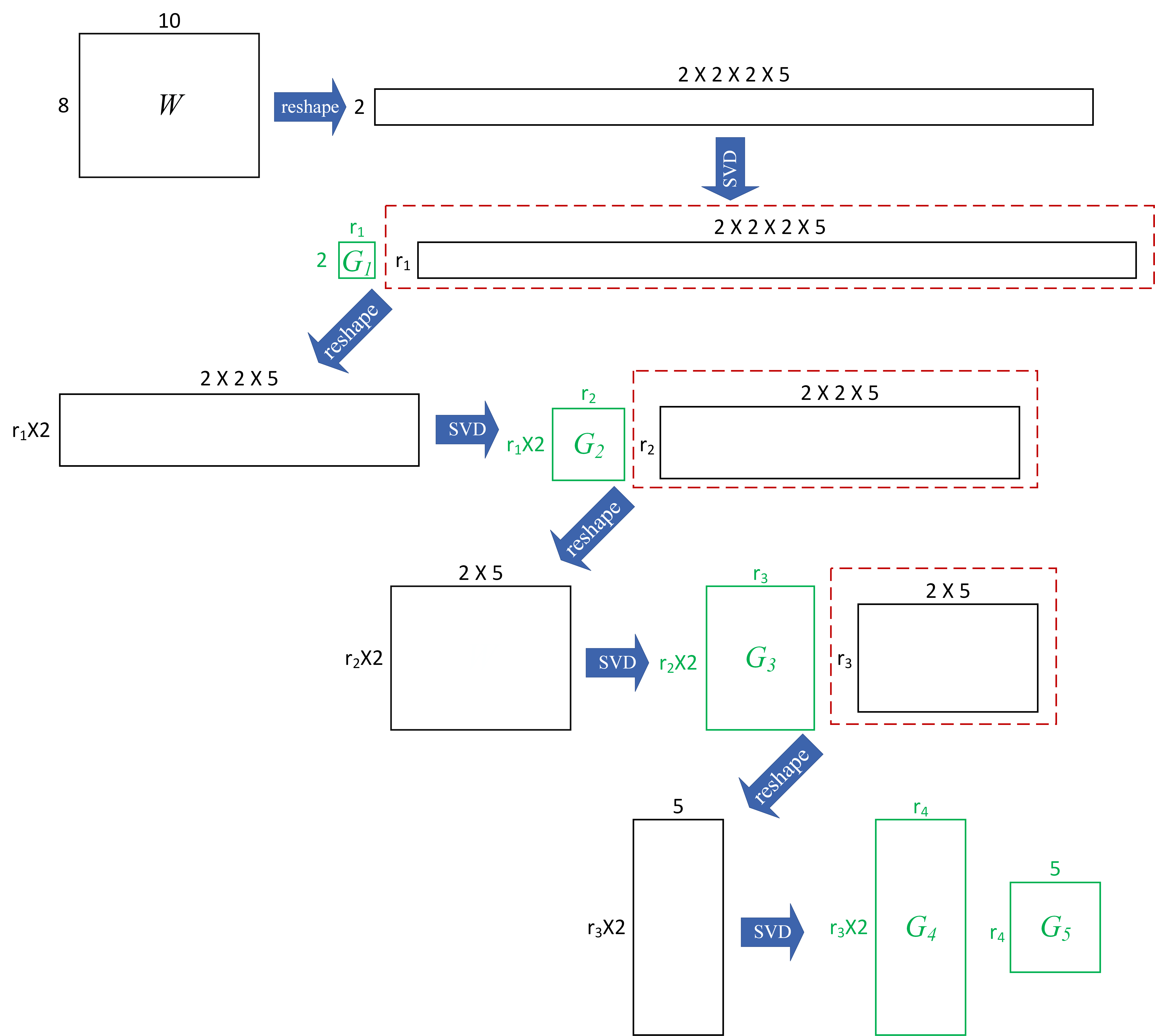}}
\caption{Tensor Train Decomposition of a matrix of size $8 \times 10$ by recursively applying reshaping and low-rank SVD. After decomposition and reshaping, the dimensions of the five TT-cores are $1 \times 2 \times r_1$, $r_1 \times 2 \times r_2$, $r_2 \times 2 \times r_3$, $r_3 \times 2 \times r_4$ and $r_4 \times 5 \times 1$, respectively.}
\label{fig:TTdet}
\end{figure*}

Tensor Train Decomposition utilizes two key ideas: recursively applying low-rank SVD and reshaping if the matrix is too thin. As illustrated in Fig.~\ref{fig:TTdet}, a matrix of size $8 \times 10$ can be decomposed into five matrices $G_1$ through $G_5$ by recursively applying reshaping and low-rank SVD. Then these five matrices can be folded into five TT-cores of sizes $1 \times 2 \times r_1$, $r_1 \times 2 \times r_2$, $r_2 \times 2 \times r_3$, $r_3 \times 2 \times r_4$ and $r_4 \times 5 \times 1$, respectively.

\subsection{Canonical Polyadic Decomposition}
In 1927, the idea of expressing a tensor as the sum of a finite number of rank-one tensors was proposed by Hitchcock~\cite{hitchcock1927expression,hitchcock1928multiple}. In 1970, the concept was generalized as CANDECOMP (canonical decomposition) by Carroll and Chang~\cite{carroll1970analysis} and as PARAFAC (parallel factors) by Harshman~\cite{harshman1970foundations}.

\begin{figure}[htbp]
\centering
\resizebox{0.485\textwidth}{!}{%
\includegraphics{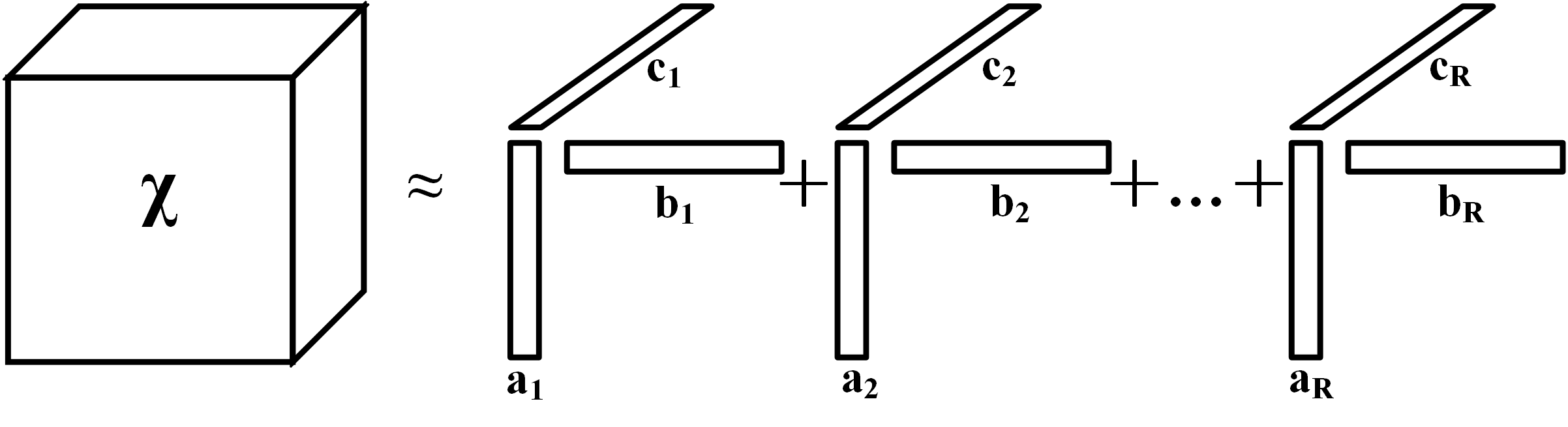}}
\caption{CP decomposition of a third-order tensor.}
\label{fig:cpd}
\end{figure}

The Canonical Polyadic (CP) decomposition factorizes a tensor into a sum of component rank-one tensors. For example, a third-order tensor $\mathcal{X} \in \mathbb{R}^{I \times J \times K}$ can be approximated as:

\begin{equation}
\centering
\mathcal{X}\approx \sum_{r=1}^R \bm{a}_r \circ \bm{b}_r \circ \bm{c_r},\nonumber
\end{equation}
where $\circ$ denotes the outer product of vectors. Parameter $R$ is a positive integer and $\bm{a}_r \in \mathbb{R}^I$, $\bm{b}_r \in \mathbb{R}^J$ and $\bm{c}_r \in \mathbb{R}^K$ for $r = 1, \dots, R$. As shown in Fig.~\ref{fig:cpd}, each element of tensor $\mathcal{X}$ can be computed as:

\begin{equation}
\centering
\bm{x}_{i,j,k}\approx \sum_{r=1}^R a_{ir} b_{jr} c_{kr},\nonumber
\end{equation}
where $i=1,\dots, I$, $j = 1,\dots,J$, $k = 1,\dots, K$. The factor matrices are defined as the combination of the vectors from the rank-one components: $\bm{A} = [\bm{a}_1\quad \bm{a}_2\quad \dots \quad \bm{a}_R]$, $\bm{B} = [\bm{b}_1\quad \bm{b}_2\quad \dots \quad \bm{b}_R]$ and $\bm{C} = [\bm{c}_1\quad \bm{c}_2\quad \dots \quad \bm{c}_R]$.

The rank of a tensor $\mathcal{X}$ is defined as the smallest number of rank-one tensors that can generate the original tensor as their sum. Determining the rank of a specifically given tensor is NP-hard~\cite{haastad1990tensor}. There is no straightforward algorithm to solve this problem. For a general third-order tensor $\mathcal{X} \in \mathbb{R}^{I \times J \times K}$, the weak upper bound on its largest attainable rank is given by:

\begin{equation}
\centering
rank(\mathcal{X})\leq min\{IJ,IK,JK\}.\nonumber
\end{equation}

Any rank that occurs with positive probability is called a typical rank. Table~\ref{table:typicalrank} from~\cite{kolda2009tensor} shows known typical ranks of specific third-order tensors over $\mathbb{R}$.

\begin{table}[htbp]
\centering
\caption{Typical ranks for third-order tensors from~\cite{kolda2009tensor}.}
\begin{tabular}{|c|c|c|}
\hline
Tensor dimension & Typical rank & Reference \\ \hline
$2\times 2\times 2$            & \{2,3\}      &\cite{kruskal1989rank}          \\ \hline
$3\times 3\times 2$            & \{3,4\}      &\cite{kruskal1983statement,ten1991kruskal}           \\ \hline
$5\times 3\times 3$            & \{5,6\}      &\cite{ten2004partial}           \\ \hline
$I\times J\times 2$ with ($I \geq 2J$) & $2J$ &\cite{ten1999simplicity}           \\ \hline
$I\times J\times 2$ with ($J < I< 2J$) & $I$  &\cite{ten1999simplicity}           \\ \hline
$I\times I\times 2$            & $\{I,I+1\}$  &\cite{ten1991kruskal,ten1999simplicity}           \\ \hline
$I\times J\times K$ with ($I \geq JK$) &$JK$  &\cite{ten2000typical}           \\ \hline
$I\times J\times K$ with ($JK - J < I < JK$)& $I$ &\cite{ten2000typical}            \\ \hline
$I\times J\times K$ with ($I = JK-J$) & $\{I,I+1\}$ &\cite{ten2000typical}            \\ \hline
\end{tabular}
\label{table:typicalrank}
\end{table}

Given the rank $R$, there are many algorithms to compute the CP decomposition. Denton~\textit{et al.} computed the CP decomposition by the alternating least squares (ALS) method~\cite{denton2014exploiting}. The ALS method was proposed in the original papers by Carroll and Chang~\cite{carroll1970analysis} and Harshman~\cite{harshman1970foundations}.

\begin{algorithm}[htbp]
\caption{ALS Algorithm~\cite{carroll1970analysis,harshman1970foundations}.}
\textbf{procedure CP-ALS($\mathcal{X},R)$}\\
\text{initialize $\bm{A}^{(n)} \in \mathbb{R}^{I_n \times R}$ for $n=1,\dots,N$}\\
\textbf{repeat}
\begin{algorithmic}
    \FOR{$n=1,\dots, N$}
        \STATE $\bm{V} \longleftarrow {\bm{A}^{(1)}}^\intercal \bm{A}^{(1)} * \dots * {\bm{A}^{(n-1)}}^\intercal \bm{A}^{(n-1)} * {\bm{A}^{(n+1)}}^\intercal \bm{A}^{(n+1)}* \dots * {\bm{A}^{(N)}}^\intercal \bm{A}^{(N)}$;\\ 
        $\bm{A}^{(n)} \longleftarrow \bm{X}^{(n)} (\bm{A}^{(N)} \odot \dots \odot \bm{A}^{(n+1)} \odot \bm{A}^{(n-1)} \odot \dots \odot \bm{A}^{(1)})\bm{V}^\dag$;\\
        normalize columns of $\bm{A}^{(n)}$ (storing norms as $\lambda$);
    \ENDFOR
\end{algorithmic}
\textbf{until} fit ceases to improve or maximum iterations exhausted\\
return $\lambda, \bm{A}^{(1)},\bm{A}^{(2)}, \dots, \bm{A}^{(N)}$\\
\textbf{end procedure}
\hrule
\label{alg:als}
\end{algorithm}

Given the rank $R$, the ALS procedure for an $N$-th order tensor is shown in Algorithm~\ref{alg:als} where $*$ and $\odot$ stand for the Hadamard and Khatri-Rao products of matrices, respectively. The factor matrices can be initialized randomly. The pseudo-inverse of a matrix $\bm{V}$ of size $R \times R$ must be computed at each iteration. The iterations stop when some stop conditions are satisfied, e.g., meeting the maximum iterations or little or no change in the factor matrices. The drawback of this algorithm is that subtracting the best rank-one tensor may increase tensor rank~\cite{stegeman2010subtracting}. In~\cite{lebedev2014speeding}, the authors used the non-linear least squares (NLS) method. Given the rank $R$, NLS minimizes the $L^2$-norm of the approximation error by using Gauss-Newton optimization. NLS decomposition has significantly higher accuracy than the ALS with or without fine-tuning~\cite{lebedev2014speeding}. The Krylov-Levenberg-Marquardt algorithm was used for the CP decomposition with bounded sensitivity constraint in~\cite{tichavsky2019sensitivity}. Furthermore, the authors in~\cite{phan2020stable} propose a variant of the Error Preserving Correction (EPC) method that minimizes the sensitivity of the decomposition:

\begin{equation}
    \begin{split}
        &\min_{\{\bm{A},\bm{B},\bm{C}\}} ss([[\bm{A}, \bm{B}, \bm{C}]])\\ \nonumber
        & \text{s,t.} ||\mathcal{K}-[[\bm{A}, \bm{B}, \bm{C}]]||_F^2 \leq \delta^2
    \end{split}
\end{equation}
where $ss([[\bm{A}, \bm{B}, \bm{C}]])$ represents the sensitivity of the CP decomposition and can be computed by~\cite{phan2020stable}:

\begin{equation}
\begin{split}
&ss([[\bm{A}, \bm{B}, \bm{C}]]) = K \cdot \tr{\{(\bm{A}^\intercal\bm{A})\circledast(\bm{B}^\intercal\bm{B})\}} + \\ \nonumber
& I \cdot \tr{\{(\bm{B}^\intercal\bm{B})\circledast(\bm{C}^\intercal\bm{C})\}} + J \cdot \tr{\{(\bm{A}^\intercal\bm{A})\circledast(\bm{C}^\intercal\bm{C})\}}, \nonumber
\end{split}
\end{equation}
where $\circledast$ denotes the Hadamard element-wise product. The boundary $\delta^2$ can be treated as the approximation error of the CP decomposition. See~\cite{phan2020stable} for further details.

% \subsection{Tensor Train Decomposition}
% Tensor Train Decomposition can be applied to the feed-forward layers of RNN models~\cite{yang2017tensor}. Given a $d$-dimensional target tensor $\mathcal{X} \in \mathbb{R}^{I_1 \times I_2 \times \cdots \times I_d}$, it can be decomposed as~\cite{yang2017tensor}:
% \begin{equation}
% \centering
% \mathcal{X} \overset{\mathrm{TTF}}{=} \mathcal{G}_1(I_1) \mathcal{G}_2(I_2) \cdots \mathcal{G}_d(I_d) \nonumber
% \end{equation}
% where
% \begin{equation}
% \begin{split}
% &\mathcal{G}_k \in \mathbb{R}^{p_k \times r_{k-1} \times r_k}, I_k \in [1, p_k] \quad \forall k \in [1, d] \\
% &\text{and } r_0 = r_d = 1. \nonumber
% \end{split}
% \end{equation}

% As illustrated in Fig.~\ref{fig:ttd}, the target tensor is represented by a sequence of matrix multiplications. Each $\mathcal{G}_k$ is a core-tensor and the complexity of the Tensor Train Decomposition is determined by the ranks ($r_0, r_1, \cdots, r_d$). Notice that the first and last ranks are restricted to be $1$ which ensures that the first and last core tensors are matrices so that the output of the chain of multiplications is a scalar.

% \begin{figure}[htbp]
% \centering
% \resizebox{0.485\textwidth}{!}{%
% \includegraphics{fig/TTd.png}}
% \caption{An illustration of Tensor Train Decomposition for a $d$-order tensor~\cite{yang2017tensor}.}
% \label{fig:ttd}
% \end{figure}

\subsection{Tucker Decomposition}
\begin{figure}[htbp]
\centering
\resizebox{0.485\textwidth}{!}{%
\includegraphics{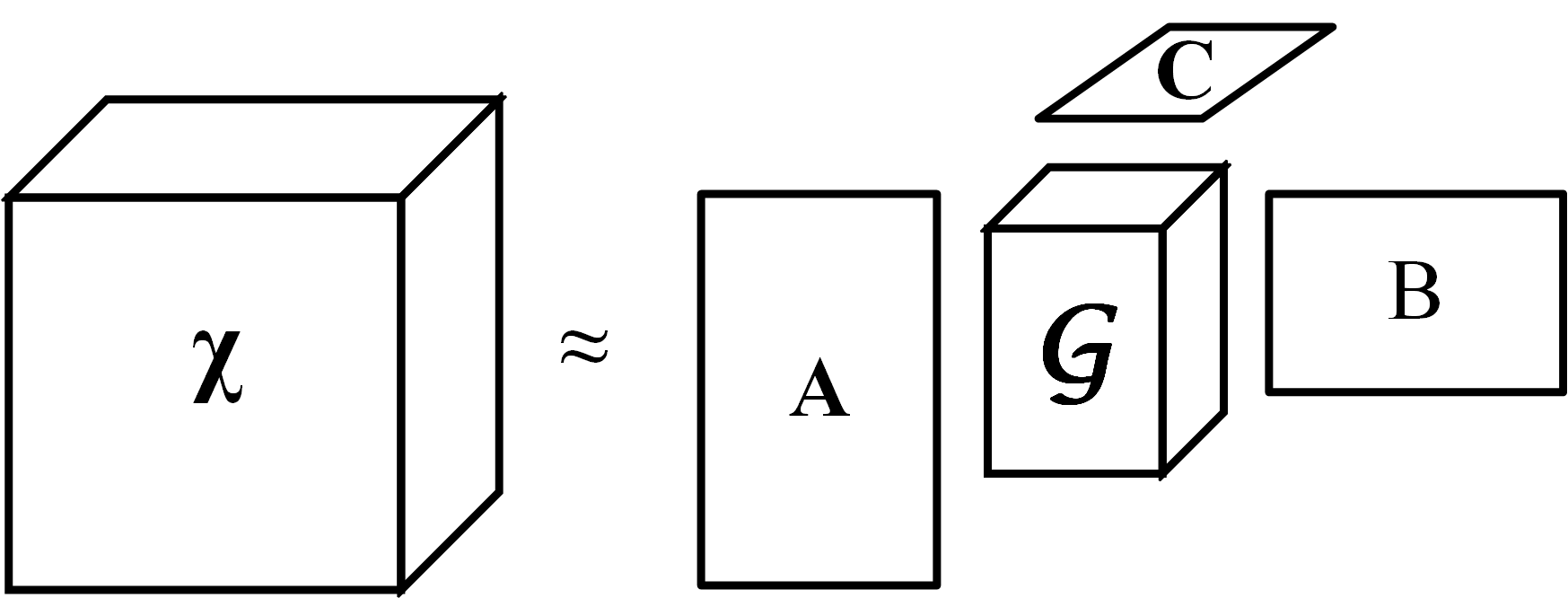}}
\caption{Tucker decomposition of a third-order tensor $\mathcal{X}$.}
\label{fig:tkd}
\end{figure}

Tucker decomposition was first proposed by Tucker~\cite{tucker1963implications,tucker1966some}. The Tucker decomposition can be treated as a form of higher-order principal component analysis (PCA). A third-order tensor $\mathcal{X} \in \mathbb{R}^{I \times J \times K}$ can be decomposed into a core tensor multiplied by a matrix along each mode:

\begin{equation}
\centering
\begin{split}
\mathcal{X} &\approx \mathcal{G} \bullet_1 \bm{A} \bullet_2 \bm{B} \bullet_3 \bm{C} \\ \nonumber
& = \sum_{p=1}^P \sum_{q=1}^Q \sum_{r=1}^R g_{pqr} \bm{a}_p \circ \bm{b}_q \circ \bm{c}_r \\ \nonumber
& = [[\mathcal{G};\bm{A}, \bm{B}, \bm{C}]], \nonumber
\end{split}
\end{equation}
where $\bm{A} \in \mathbb{R}^{I \times P}$, $\bm{B} \in \mathbb{R}^{J \times Q}$ and $\bm{C} \in \mathbb{R}^{K \times R}$ are the factor matrices and can be treated as the principal components in each mode. The tensor $\mathcal{G} \in \mathbb{R}^{P \times Q \times R}$ is the core tensor. In the Tucker decomposition, each element of the original tensor $\mathcal{X}$ can be represented by:

\begin{equation}
\centering
x_{i,j,k} \approx \sum_{p=1}^P \sum_{q=1}^Q \sum_{r=1}^R g_{pqr} a_{ip} b_{jq} c_{kr},
\end{equation}
for $i=1, \dots, I$, $j=1, \dots, J$ and $k=1, \dots, K$. The parameters $P$, $Q$, and $R$ represent the number of components in the factor matrices $\bm{A}$, $\bm{B}$ and $\bm{C}$, respectively. Fig.~\ref{fig:tkd} shows the Tucker decomposition of a third-order tensor. In fact, CP decomposition can be viewed as a special case of the Tucker decomposition where the core tensor is superdiagonal and $P$, $Q$ and $R$ are identical.

\subsection{Tensor Ring Decomposition}
The limitation of setting the ranks hampers the flexibility and representation capability of the Tensor Train based models. When multiplying TT-cores, a strict order must be followed and the alignment of the tensor dimensions is critical in obtaining the optimized TT-cores. However, locating the best alignment is a hard problem~\cite{zhao2016tensor}.

The main modification of the tensor ring decomposition (TRD) is connecting the first and the last core tensors circularly to build a ring-like structure~\cite{pan2019compressing}. It can alleviate the limitation of the tensor train models as mentioned above. Formally, a $d$-dimensional target tensor $\mathcal{X} \in \mathbb{R}^{I_1 \times I_2 \times \cdots \times I_d}$ can be decomposed as~\cite{pan2019compressing}:
\begin{equation}
\centering
\mathcal{X} \overset{\mathrm{TRD}}{=} \sum_{r_0 = r_d, r_2, \cdots, r_{d-1}} \mathcal{G}^{(1)}_{r_0, I_1, r_1}\mathcal{G}^{(2)}_{r_1, I_2, r_2}\cdots \mathcal{G}^{(d)}_{r_{d-1}, I_d, r_d}\nonumber
\end{equation}

\begin{figure}[htbp]
\centering
\resizebox{0.25\textwidth}{!}{%
\includegraphics{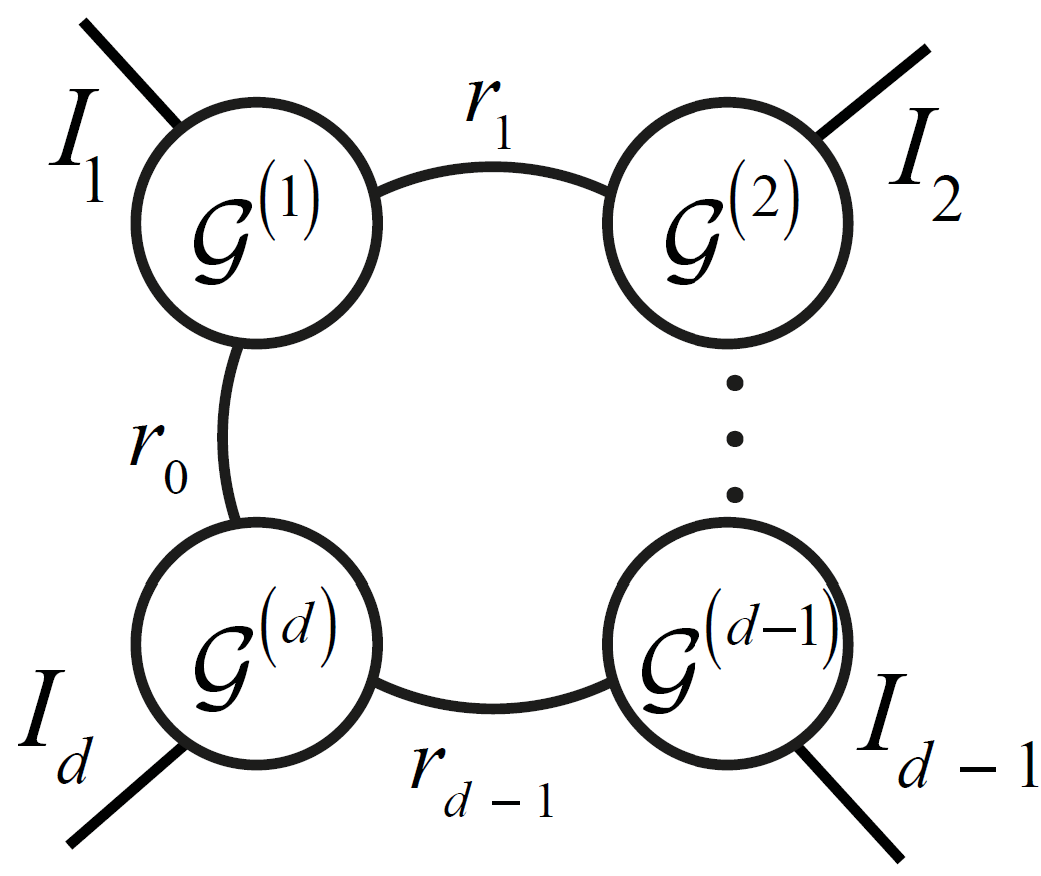}}
\caption{Tensor Ring Decomposition in a ring form: the core tensors are multiplied one by one and form a ring~\cite{pan2019compressing}.}
\label{fig:trd}
\end{figure}

The first and the last ranks are set to $R$ ($R > 1$). For any index $k$ ($k \in {1, 2, \cdots, R}$), the first order of the first core tensor $\mathcal{G}^{(1)}_{r_0=k}$ and the last order of the last core tensor $\mathcal{G}^{(d)}_{r_d=k}$ are matrices. The tensor ring structure can be regarded as a summation of $R$ of tensor trains along each of the $R$ slices of tensor cores. The product $\mathcal{G}^{(1)}_{k,I_1} \mathcal{G}^{(2)}_{I_2} \cdots \mathcal{G}^{(d)}_{I_d, k}$ has the form of tensor train decomposition by fixing $r_0=r_d=k$. Therefore, a tensor ring model is a linear combination of $R$ different tensor train models. The tensor ring structure is illustrated in Fig.~\ref{fig:trd}.

\subsection{Block Term Decomposition}
\begin{figure}[htbp]
\centering
\resizebox{0.485\textwidth}{!}{%
\includegraphics{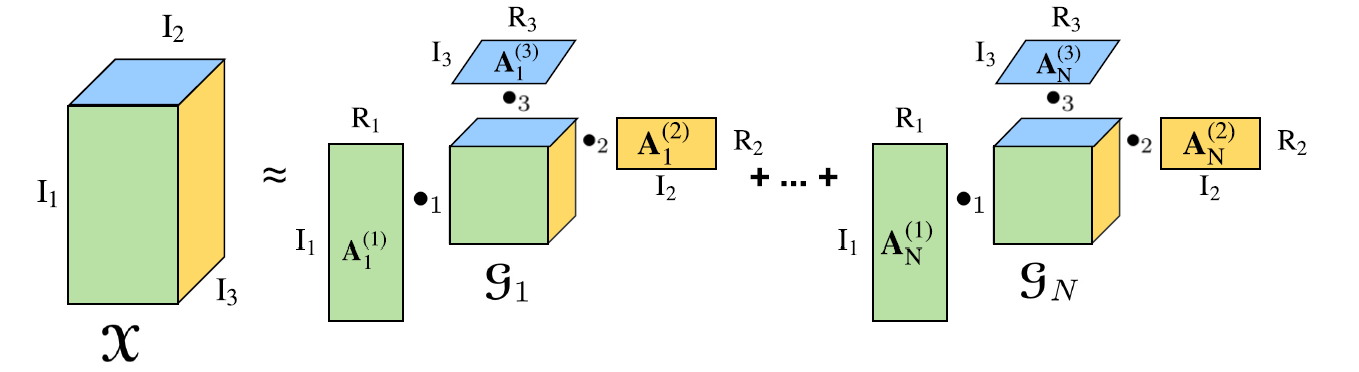}}
\caption{Block Term decomposition for a $3^{rd}$-order tensor. The tensor can be approximated by $N$ Tucker decompositions where $N$ is the CP-rank; $R_1, R_2, R_3$ are the Tucker-rank; d is the Core-order~\cite{ye2018learning}.}
\label{fig:btd}
\end{figure}

Block Term decomposition combines CP decomposition and Tucker decomposition~\cite{ye2018learning}. Given a $d$-dimensional target tensor $\mathcal{X} \in \mathbb{R}^{I_1 \times I_2 \times \cdots \times I_d}$, it can be decomposed into $N$ block terms as follows~\cite{ye2018learning}:
\begin{equation}
\centering
\mathcal{X} \overset{\mathrm{BTD}}{=} \sum_{n=1}^N \mathcal{G}_n \bullet_1 \mathcal{A}_n^{(1)} \bullet_2 \mathcal{A}_n^{(2)} \bullet_3 \cdots \bullet_d \mathcal{A}^{(d)}_n \nonumber
\end{equation}
where each term computes tensor-tensor product on $k^{th}$ order ($\bullet_k$) between a core tensor $\mathcal{G}_n \in \mathbb{R}^{R_1 \times \cdots \times R_d}$ and $d$ factor matrices $\mathcal{A}^{(k)}_n \in \mathbb{R}^{I_k \times R_k}$ of $\mathcal{G}_n$'s $k^{th}$ dimension, where $n \in [1, N]$ and $k \in [1, d]$~\cite{de2008decompositions}. $N$ is the CP-rank. $R_1, R_2, \cdots, R_d$ are the Tucker-rank where $d$ is the Core-order. Fig.~\ref{fig:btd} shows the block term decomposition for a $3^{rd}$-order tensor.

\subsection{Hierarchical Tucker Decomposition}
\begin{figure}[htbp]
\centering
\resizebox{0.485\textwidth}{!}{%
\includegraphics{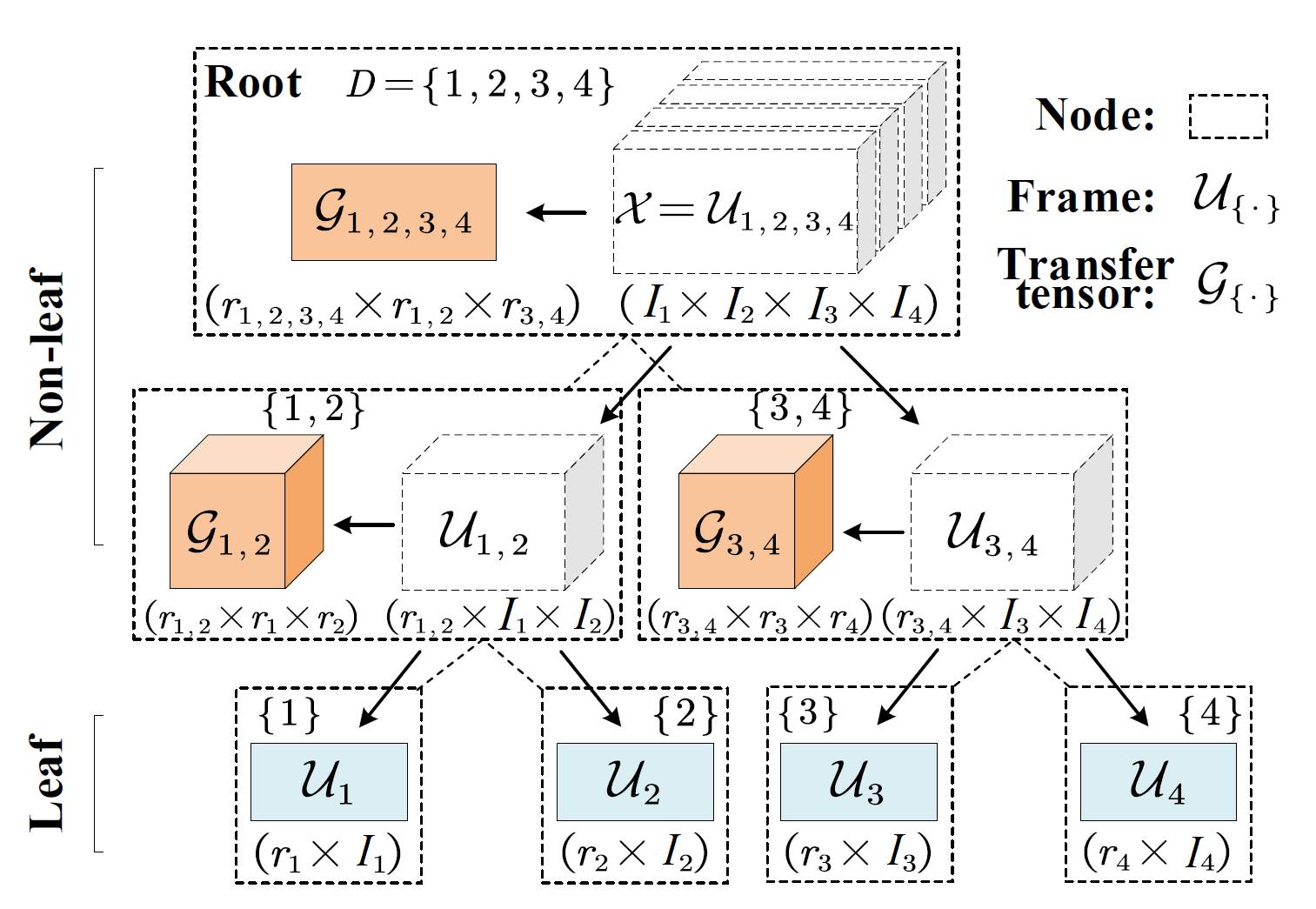}}
\caption{Hierarchical Tucker decomposition for a $4^{th}$-order tensor. It is a binary tree with root $D = \{1, 2, 3, 4 \}$ where the dashed boxes represent the nodes. Node $\{1\}$ is a leaf node whose parent and sibling are node $\{1, 2\}$ and node $\{2\}$, respectively~\cite{yin2020compressing}.}
\label{fig:htd}
\end{figure}

Hierarchical Tucker decomposition has multiple hierarchical levels based on the order of the tensor. From top to bottom in a binary tree, a hierarchical tucker decomposed tensor can be recursively decomposed into intermediate tensors, referred as \textit{frames}, as shown in Fig.~\ref{fig:htd}. Each frame is a unique \textit{node} where each node is associated with a \textit{dimension set}. Given a $d$-dimensional target tensor $\mathcal{X} \in \mathbb{R}^{I_1 \times I_2 \times \cdots \times I_d}$, a binary tree with a root node associated with $D=\{1, 2, \cdots, d\}$ can be built where $\mathcal{X} = \mathcal{U}_D$ is the root frame. For each non-leaf frame $\mathcal{U}_s \in \mathbb{R}^{r_s \times I_{\mu_s} \times \cdots \times I_{\nu_s}}$, $s \subsetneq D$ is associated with the node corresponding to $\mathcal{U}_s$ and $s_1, s_2 \subsetneq s$ are associated with the left and right child nodes of the $s$-associated node where $\mu_s = min(s)$ and $\nu_s = max(s)$. A non-leaf frame $\mathcal{U}_s$ can be recursively decomposed to a left child frame ($\mathcal{U}_{s_1}$), a right child frame ($\mathcal{U}_{s_2}$) and a transfer tensor (($\mathcal{G} \in \mathbb{R}^{r_s \times r_{s_1} \times r_{s_2}}$)) as follows\cite{yin2020compressing}:
\begin{equation}
\centering
\mathcal{U}_s = \mathcal{G}_s \times_1^2 \mathcal{U}_{s_1} \times_1^2 \mathcal{U}_{s_2}, \nonumber
\end{equation}
where $\times_1^2$ denotes the tensor contraction that can be executed between two tensors with at least one matched dimension. For example, given two tensors $\mathcal{A} \in \mathbb{R}^{n_1 \times n_2 \times l}$ and $\mathcal{B} \in \mathbb{R}^{l \times m_1 \times m_2}$ where the third dimension of $\mathcal{A}$ matches the first dimension of $\mathcal{B}$, a tensor of size $n_1 \times n_2 \times m_1 \times m_2$ can be computed by using the tensor contraction operation as $(\mathcal{A} \times_1^3 \mathcal{B})_{i_1, i_2, j_1, j_2} = \sum_{\alpha = 1}^l \mathcal{A}_{i_1, i_2, \alpha} \mathcal{B}_{\alpha, j_1, j_2}$.
The original $I_1 \times \cdots \times I_d$-order tensor $\mathcal{X} = \mathcal{U}_D$ can be recursively decomposed into a combination of the $2^{nd}$-order leaf frames and the $3^{rd}$-order transfer tensors by performing the hierarchical tucker decomposition from top to bottom of the binary tree. The parameter $r_s$ is defined as \textit{hierarchical rank}.

\section{Tensorizing Convolutional Neural Networks}
\label{section:cnn}

\begin{figure}[htbp]
\centering
\resizebox{0.38\textwidth}{!}{%
\includegraphics{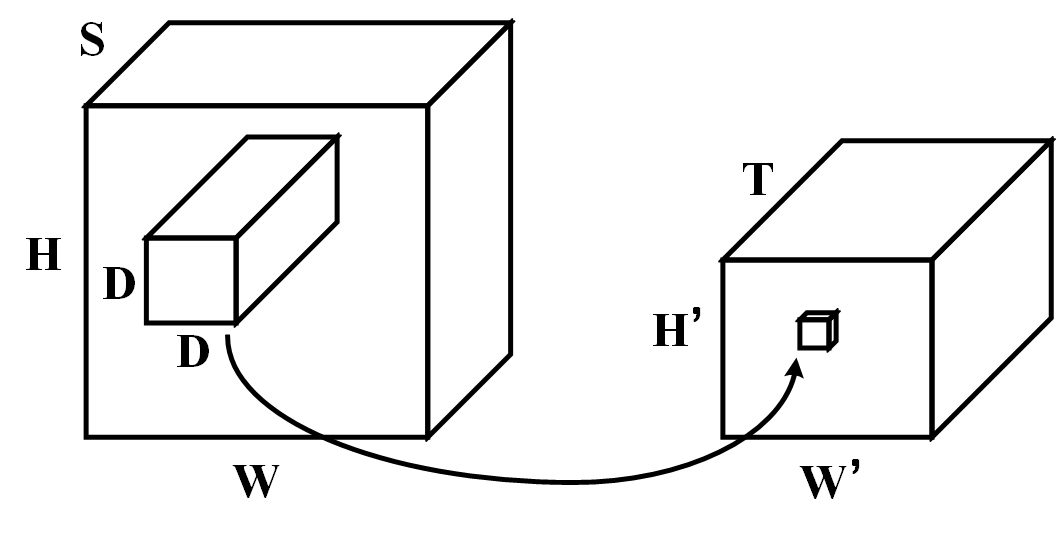}}
\caption{Original convolution layer with filter size of $T \times S \times D \times D$.}
\label{fig:conv}
\end{figure}

CNNs have two main parts: convolutional layers and fully connected layers. In general, convolutional layers in CNNs map a third-order input tensor $\mathcal{X}$ of size $S \times W \times H$ into a third-order output tensor $\mathcal{Y}$ of size $T \times W' \times H'$ with a $4^{th}$-order kernel tensor $\mathcal{K}$ of size $T \times S \times D \times D$, where $D \times D$ represents the filter size, $S$ and $T$ represent the number of input and output channels, respectively. The typical convolutional filter sizes are small, e.g., $3 \times 3$, $7 \times 7$, compared to the numbers of input ($S$) and output ($T$) channels. Each convolutional layer may have hundreds or thousands of filters which are suited for tensor decomposition methods. As shown in Fig.~\ref{fig:conv}, the output tensor $\mathcal{Y}$ can be computed as:

\begin{equation}
\centering
\mathcal{Y}_{t,w',h'} = \sum_{s=1}^S \sum_{j=1}^D \sum_{i=1}^D \mathcal{K}_{t,s,j,i}\mathcal{X}_{s,w_j,h_i},\nonumber
\end{equation}
where $w_j=(w'-1)q+j-p$ and $h_i=(h'-1)q+i-p$. The parameters $p$ and $q$ represent zero-padding size and stride, respectively.

Fully connected layers apply a linear transformation to an $N$-dimensional input vector $\bm{x}$ and compute a $M$-dimensional output vector $\bm{y}$:

\begin{equation}
\centering
\bm{y}=\bm{W}\bm{x}+\bm{b}.
\label{eq:fc}
\end{equation}
where $\bm{W}$ and $\bm{b}$ represent the weight matrix and the bias vector, respectively. In this section, we show how these different tensor decomposition methods can be applied to CNNs.

\subsection{Tensor Train Decomposition}
Tensor Train Decomposition can be applied to both the convolutional layers and the fully connected layers of CNNs. Let's first describe the implementation of TT-decomposition on the fully connected layers.

Now consider the TT-representation of the weight matrix $\bm{W} \in \mathbb{R}^{M \times N}$ where $M = \prod_{k=1}^d m_k$ and $N = \prod_{k=1}^d n_k$. Define two bijection mappings $\mu(t)=(\mu_1(t), \mu_2(t), \dots, \mu_d(t))$ and $\nu(t)=(\nu_1(t), \nu_2(t), \dots, \nu_d(t))$ that map the matrix $\bm{W}$ of size $M\times N$ to a higher-order tensor $\mathcal{W}$ of size $n_1m_1\times n_2m_2\times \dots n_dm_d$. The mapping $\mu(\cdot)$ maps the row index $\ell = 1, 2, \dots, M$ of the matrix $\bm{W}$ into a $d$-dimensional vector-indices whose $k$-th dimensions are of length $m_k$. The mapping $\nu(\cdot)$ maps the column index $t = 1, 2, \dots, N$ of the matrix $\bm{W}$ into $d$-dimensional vector-indices whose $k$-th dimensions are of length $n_k$. Thus the $k$-th dimension of the reshaped $d$-dimensional tensor $\mathcal{W}$ is of length $n_km_k$ and is indexed by the tuple ($\mu_k(\cdot), \nu_k(\cdot)$). Then the tensor $\mathcal{W}$ can be converted using TT-decomposition:

\begin{equation}
\begin{split}
\bm{W}(\ell,t) &= \mathcal{W}((\mu_1(\ell), \nu_1(t)),\dots,(\mu_d(\ell), \nu_d(t))) \\ \nonumber
&=\bm{G}_1[(\mu_1(\ell), \nu_1(t))]\dots \bm{G}_d[(\mu_d(\ell), \nu_d(t))].
\end{split}
\end{equation}

\begin{figure*}[htbp]
\centering
\resizebox{0.9\textwidth}{!}{%
\includegraphics{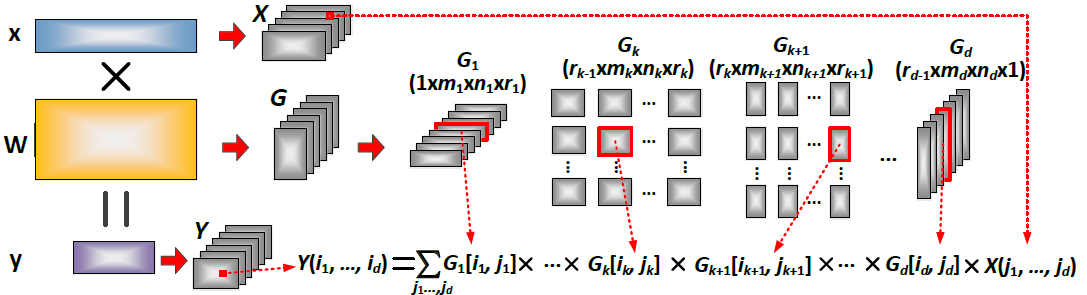}}
\caption{TT-format fully connected layer~\cite{deng2019tie}.}
\label{fig:TTfc}
\end{figure*}

The TT-format of the weight matrix transforms a $d$-dimensional tensor $\mathcal{X}$ (formed from the input vector $\bm{x}$) to the $d$-dimensional tensor $\mathcal{Y}$ (which can be used to compute output vector $\bm{y}$). As illustrated in Fig.~\ref{fig:TTfc} from~\cite{deng2019tie}, the linear transformation of a fully connected layer can be computed in the TT-format:

\begin{equation}
\begin{split}
\mathcal{Y}(i_1, \dots, i_d) =& \sum_{j_1,\dots,j_d} \bm{G}_1[i_1, j_1]\dots \bm{G}_d[i_d, j_d]\mathcal{X}(j_1, \dots, j_d) \\ \nonumber
 &+\mathcal{B}(i_1, \dots, i_d).\nonumber
\end{split}
\end{equation}
where $\mathcal{B}$ corresponds to the bias vector $\bm{b}$ in Eq.~(\ref{eq:fc}). The ranks of the TT-format for the weight matrix depend on the choice of the bijection mappings $\mu(t)$ and $\nu(t)$. The computational complexity of the forward pass is $\mathcal{O}(dr^2m\text{max}\{M,N\})$.

In machine learning, backpropagation is widely used to train feed-forward neural networks based on the stochastic gradient descent algorithm~\cite{rumelhart1986learning}. Backpropagation computes the gradient of the loss-function $L$ with respect to all the parameters. Given the gradients with respect to  the layer's output $\frac{\partial L}{\partial \bm{y}}$, the backpropagation applied to the fully connected layers computes the gradients with respect to the input $\bm{x}$, the weight matrix $\bm{W}$ and bias vector $\bm{b}$:

\begin{equation}
\centering
\frac{\partial L}{\partial \bm{x}}= \bm{W}^\intercal\frac{\partial L}{\partial \bm{y}},\frac{\partial L}{\partial \bm{W}}=\frac{\partial L}{\partial \bm{y}}\bm{x}^\intercal,\frac{\partial L}{\partial \bm{b}}=\frac{\partial L}{\partial \bm{y}}.\nonumber
\end{equation}

Notice that the gradient of the loss function with respect to the bias vector is the same as that with respect to the output. The gradient of the loss function with respect to the input vector $\bm{x}$ can be computed using the same matrix-by-vector product as shown in Fig.~\ref{fig:TTfc} with the complexity of $\mathcal{O}(dr^2m\text{max}\{M,N\})$. Instead of directly computing $\frac{\partial L}{\partial \bm{W}}$ which requires $\mathcal{O}(MN)$ memory, it's better to compute the gradient of the loss function $L$ with respect to each TT-cores. The overall computation complexity of the backpropagation in the fully connected layer is $\mathcal{O}(d^2r^4m\text{max}\{M,N\})$~\cite{novikov2015tensorizing}. See~\cite{novikov2015tensorizing} for a detailed description of all the learning processes in the fully connected layers in TT-format.

We now review the implementation of TT-decomposition on convolutional layers. One straightforward way to represent the convolutional kernel $\mathcal{K}$ in the TT-format is to apply the TT-decomposition directly to the tensor $\mathcal{K}$. But in~\cite{garipov2016ultimate}, the disadvantage of this method was mentioned. The kernel of a $1 \times 1$ convolution is a 2-dimensional array whose TT-format coincides with the matrix low-rank format. However, the matrix TT-format is more efficient than the matrix low-rank format~\cite{novikov2015tensorizing}.

\begin{figure*}[htbp]
\centering
\resizebox{0.7\textwidth}{!}{%
\includegraphics{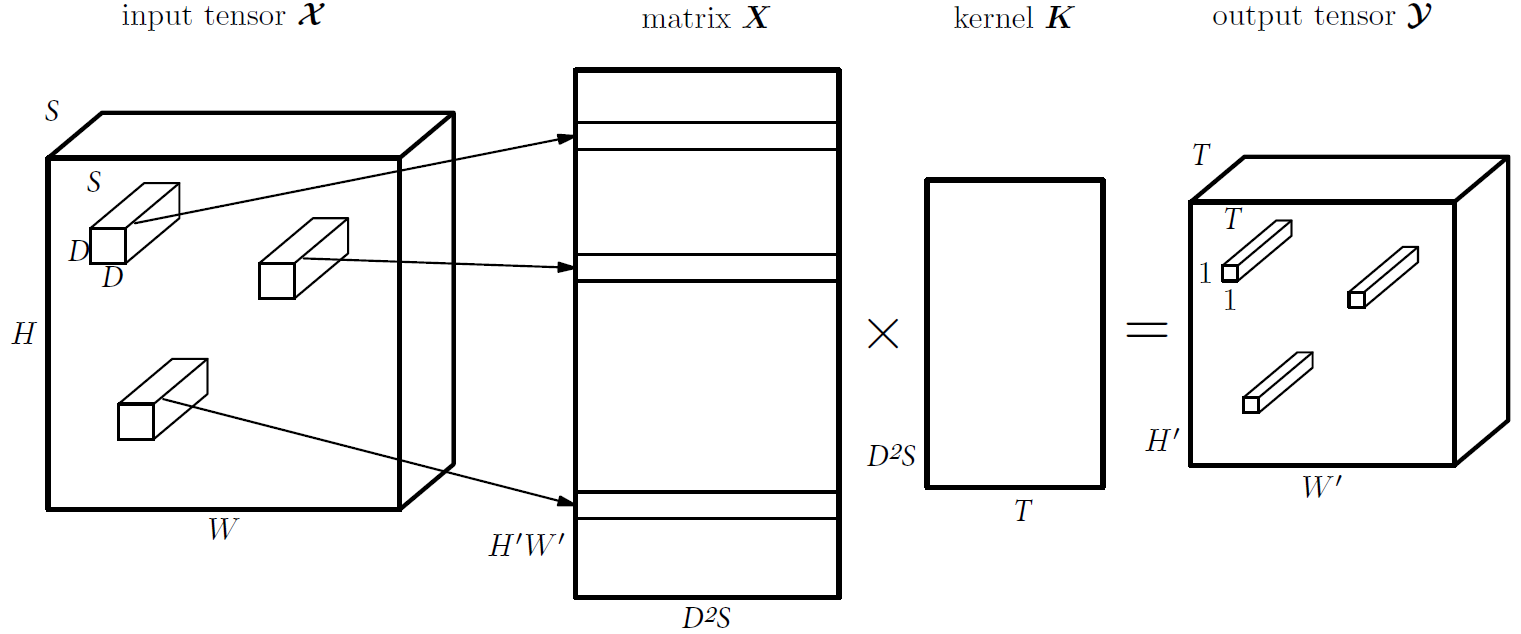}}
\caption{Reducing convolution to a matrix-by-matrix multiplication~\cite{garipov2016ultimate}.}
\label{fig:reshape}
\end{figure*}

The second approach of applying TT-decomposition to the convolutional layers is inspired by the theory that a convolutional layer can be formulated as a matrix-by-matrix multiplication~\cite{vedaldi2014convolutional,vedaldi2015matconvnet,garipov2016ultimate}. As illustrated in Fig.~\ref{fig:reshape}, the two-dimensional convolution between a three-dimensional input tensor and a four-dimensional model is equivalent to the matrix-by-matrix multiplication~\cite{garipov2016ultimate}.

In Fig.~\ref{fig:reshape}, $H'=H-D+1$ and $W'=W-D+1$, and the output tensor $\mathcal{Y} \in \mathbb{R}^{T \times W' \times H'}$ is reshaped into a matrix $\bm{Y} \in \mathbb{R}^{T \times W'H'}$ as follows:

\begin{equation}
\centering
\mathcal{Y}(t, y, x) = \bm{Y}(t, y + W'(x-1)).\nonumber
\end{equation}

The matrix $\bm{X}$ of size $W'H' \times D^2S$ whose $k$-th row represents the $S\times D \times D$ patch of the input tensor is generated by:

\begin{equation}
\centering
\resizebox{1\columnwidth}{!}{
$\mathcal{X}(s, y+j-1, x+i-1) = \bm{X}(x+W'(y-1),j+D(i-1)+D^2(S-1)),$\nonumber
}
\end{equation}
where $y=1, \dots, H'$, $x=1, \dots, W'$, $i,j=1, \dots, D$. Finally, the kernel tensor $\mathcal{K}$ can be reshaped into a matrix $\bm{K}$ of size $D^2S \times T$:

\begin{equation}
\centering
\mathcal{K}(t, s, j, i) = \bm{K}(t, j+D(i-1)+D^2(S-1)).\nonumber
\end{equation}

After all these reshapings, the convolutional layers can be written as matrix-by-vector multiplication as shown in Fig.~\ref{fig:reshape}. Then the matrix TT-format can be directly applied to the matrix $\bm{K}$: reshape it into a tensor $\mathcal{K}$ and then convert it into the TT-format. Assume that the dimension of the matrix $\bm{K}$ can be factorized as: $S = \prod_{i=1}^d S_i$ and $T = \prod_{i=1}^d T_i$. This assumption is always true since we can add some dummy channels to increase the values of $S$ and $T$. Then the ($d+1$)-dimensional convolutional kernel whose $k$-th dimension has length $D^2$ for $k = 0$ and $S_k T_k$ for $k=1,\dots,d$ can be defined as:

\begin{equation}
\begin{split}
\centering
&\bm{K}(t',x+D(y-1)+D^2(s'-1))\\ \nonumber
&=\mathcal{K}((1,x+D(y-1)),(t_1,s_1),\dots,(t_d,s_d))\\\nonumber
&=\bm{G}_0[1,x+D(y-1)]\bm{G}_1[t_1,s_1]\dots \bm{G}_d[t_d,s_d], \nonumber
\end{split}
\end{equation}
where $s'=s_1+\sum_{i=2}^d(s_i-1)\prod_{j=1}^{i-1}S_j$ and $t'=t_1+\sum_{i=2}^d(t_i-1)\prod_{j=1}^{i-1}T_j$.

First, the TT-convolution layer reshapes the input tensor into a ($d+2$)-dimensional tensor $\mathcal{X} \in \mathbb{R}^{S_1 \times \dots \times S_d \times W \times H}$ and then transforms this tensor into the output tensor $\mathcal{Y} \in \mathbb{R}^{T_1 \times \dots \times T_d \times (W-D+1) \times (H-D+1)}$ using the following equation:

\begin{equation}
\begin{split}
\centering
&\mathcal{Y}(t_1,\dots,t_d,y,x)\\ \nonumber
=&\sum_{i=1}^D \sum_{j=1}^D \sum_{s_1,\dots,s_d}\mathcal{X}(s_1,\dots,s_d,j+y-1,i+x-1)\times\\ \nonumber
&\bm{G}_0[1,y+D(x-1)]\bm{G}_1[t_1,s_1]\dots \bm{G}_d[t_d,s_d]. \nonumber
\end{split}
\end{equation}

While training the network, the stochastic gradient is applied to each element of the TT-cores with momentum~\cite{garipov2016ultimate}.

\subsection{Canonical Polyadic Decomposition}
In CNNs, convolution layers map an input tensor of size $S\ \times W \times H$ into an output tensor of size $T \times W' \times H'$ using a kernel tensor of size $T \times S \times D \times D$ where $S$ and $T$ represent different input and output channels, respectively. Now consider how to approximate the kernel tensor $\mathcal{K}$ with rank-$R$ CP-decomposition. The number of the parameters needed to represent the kernel tensor of size $T \times S \times D \times D$ after the decomposition is $R(D^2+T+S)$ since the $4$-dimensional kernels are reshaped to $3$-dimensional kernels of size $T \times S \times D^2$ as the filter size $D$ is relatively small (e.g., $3 \times 3$ or $5 \times 5$). The rank-$R$ CP format of the reshaped kernel tensor can be represented as~\cite{astrid2017cp}:

\begin{equation}
\centering
\mathcal{K}_{t,s,j,i} = \sum_{r=1}^R \bm{U}_{r,s}^{(1)}\mathcal{U}_{r,j,i}^{(2)}\bm{U}_{t,r}^{(3)},\nonumber
\end{equation}
where $\bm{U}_{r,s}^{(1)}$, $\mathcal{U}_{r,j,i}^{(2)}$,
$\bm{U}_{t,r}^{(3)}$ are the three tensors of sizes $R \times S$, $R\times D \times D$ and $T \times R$, respectively. Then the approximate transformation of the convolution from the input tensor $\mathcal{X}$ to the output tensor $\mathcal{Y}$ is given by:

\begin{equation}
\centering
\mathcal{Y}_{t,w',h'} = \sum_{r=1}^R \bm{U}_{t,r}^{(3)}(\sum_{j=1}^D \sum_{i=1}^D \mathcal{U}_{r,j,i}^{(2)}(\sum_{s=1}^S \bm{U}_{r,s}^{(1)}\mathcal{X}_{s,w_j,h_i})).\nonumber
\end{equation}

\begin{figure}[htbp]
\centering
\resizebox{0.485\textwidth}{!}{%
\includegraphics{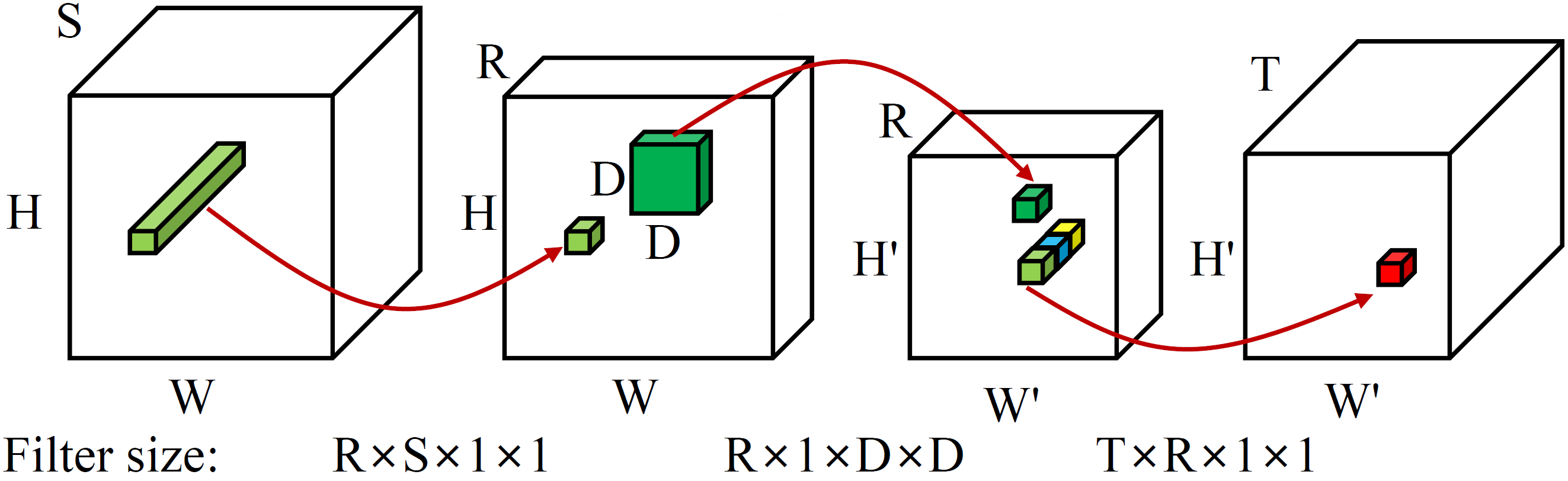}}
\caption{CP decomposed convolution layer in~\cite{astrid2017cp}.}
\label{fig:cpd3}
\end{figure}

As shown in Fig.~\ref{fig:cpd3} from~\cite{astrid2017cp}, this transformation is equivalent to a sequence of three separate small convolutional kernels in a row:

\begin{equation}
\begin{split}
\mathcal{Z}_{r,w,h}&=\sum_{s=1}^S \bm{U}_{r,s}^{(1)}\mathcal{X}_{s,w,h}, \\\nonumber
\mathcal{Z'}_{r,w',h'}& = \sum_{j=1}^D \sum_{i=1}^D \mathcal{U}_{r,j,i}^{(2)}\mathcal{Z}_{t,w_j,h_i}, \\ \nonumber
\mathcal{Y}_{t,w',h'} &= \sum_{r=1}^R \bm{U}_{t,r}^{(3)} \mathcal{Z'}_{r,w',h'}. \nonumber
\end{split}
\end{equation}

where $\mathcal{Z}_{r,w,h}$ and $\mathcal{Z'}_{r,w',h'}$ are intermediate feature tensors of sizes $R \times W \times H$ and $R\times W' \times H'$, respectively.

After this replacement, the entire network can be trained using a standard backpropagation process. Ranks play an important role in CP decomposition. Unfortunately, no polynomial time algorithm for determining the rank of a tensor exists~\cite{kruskal1989rank,haastad1990tensor}. Therefore, most algorithms approximate the tensor with different ranks until a "good" rank is found. In~\cite{astrid2017cp}, the authors applied rank-$5$ CP decomposition first and then fine-tuned the whole network to balance the accuracy loss and the number of ranks. In~\cite{phan2020stable}, authors applied a heuristic binary search to find the smallest rank such that the accuracy drop after every single layer is acceptable.

\subsection{Tucker Decomposition}
The $4^{th}$-order kernel tensor $\mathcal{K}$ can be decomposed by the rank-($R_1$, $R_2$, $R_3$, $R_4$) Tucker decomposition as:

\begin{equation}
\centering
\resizebox{1\columnwidth}{!}{
$\mathcal{K}_{t,s,j,i} = \sum_{r_1 = 1}^{R_1} \sum_{r_2 = 1}^{R_2} \sum_{r_3 = 1}^{R_3} \sum_{r_4 = 1}^{R_4} \mathcal{G'}_{r_4, r_3, r_2, r_1} \bm{U}_{r_1,i}^{(1)} \bm{U}_{r_2,j}^{(2)} \bm{U}_{r_3,s}^{(3)} \bm{U}_{r_4,t}^{(4)},$ \nonumber
}
\end{equation}
where $\mathcal{G'}$ is the core tensor of size $R_4 \times R_3 \times R_2 \times R_1$ and $\bm{U}^{(1)}$, $\bm{U}^{(2)}$, $\bm{U}^{(3)}$ and $\bm{U}^{(4)}$ are the factor matrices of sizes of sizes $R_1 \times D$, $R_2 \times D$, $R_3 \times S$ and $R_4 \times T$, respectively.

\begin{figure}[htbp]
\centering
\resizebox{0.485\textwidth}{!}{%
\includegraphics{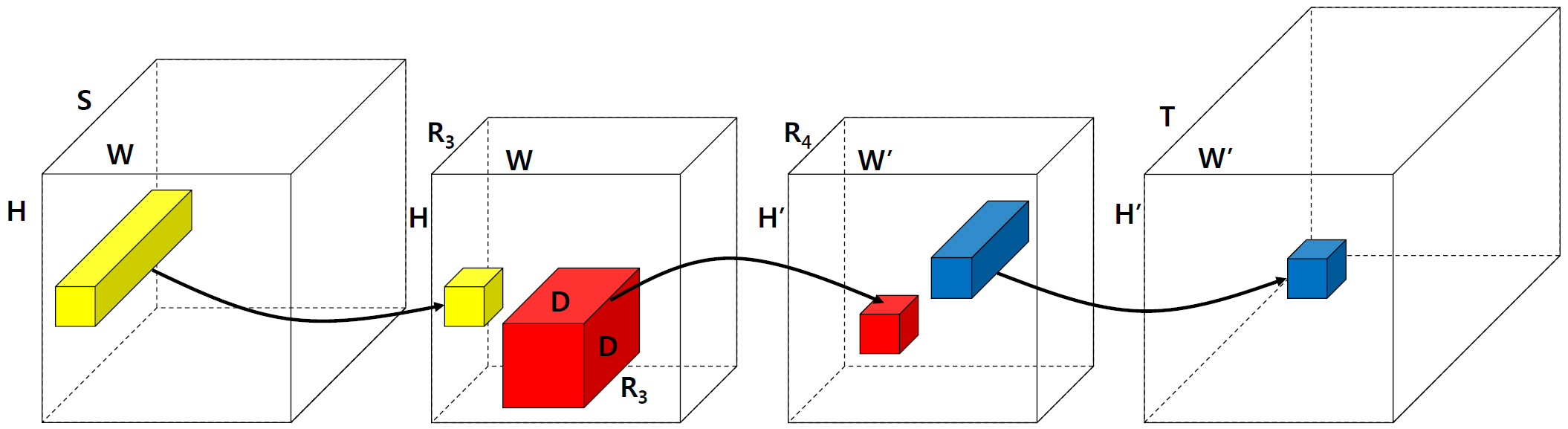}}
\caption{Tucker-2 decomposed convolution layer in~\cite{kim2015compression}.}
\label{fig:tkd3}
\end{figure}

As mentioned before, the filter size $D$ is relatively small compared to the number of input and output channels. Mode-1 and mode-2 which are associated with the filter sizes don't need to be decomposed. Under this condition, the kernel tensor can be decomposed by the Tucker-2 decomposition as follows:

\begin{equation}
\centering
\mathcal{K}_{t,s,j,i} = \sum_{r_3 = 1}^{R_3} \sum_{r_4 = 1}^{R_4} \mathcal{C}_{r_4,r_3,j,i} \bm{U}_{r_3,s}^{(3)} \bm{U}_{r_4,t}^{(4)}, \nonumber
\end{equation}
where $\mathcal{C}$ represents the core tensor of size $R_4 \times R_3 \times D \times D$. As shown in Fig.~\ref{fig:tkd3} from~\cite{kim2015compression}, this transformation is equivalent to a sequence of three separate small convolutional kernels:

\begin{equation}
\begin{split}
\mathcal{Z}_{r_3,w,h}&=\sum_{s=1}^S \bm{U}_{r_3,s}^{(3)}\mathcal{X}_{s,w,h}, \\\nonumber
\mathcal{Z'}_{r_4,w',h'}& = \sum_{r_3=1}^{R_3} \sum_{j=1}^D \sum_{i=1}^D \mathcal{C}_{r_4,r_3,j,i}\mathcal{Z}_{r_3,w_j,h_i}, \\ \nonumber
\mathcal{Y}_{t,w',h'} &= \sum_{r_4=1}^{R_4} \bm{U}_{t,r_4}^{(4)} \mathcal{Z'}_{r_4,w',h'}, \nonumber
\end{split}
\end{equation}
where $\mathcal{Z}$ and $\mathcal{Z'}$ are the intermediate tensors of sizes $R_3 \times W \times H$ and $R_4 \times W' \times H'$, respectively.

The rank-($R_3, R_4$) parameters are very important in the Tucker decomposition. They control the balance between the model compression ratio and accuracy loss. In~\cite{kim2015compression}, the authors applied global analytic solutions for variational Bayesian matrix factorization (VBMF) from~\cite{nakajima2013global} to determine the ranks of the Tucker decomposition. VBMF can be used to find noise variance, ranks and theoretical conditions for perfect rank recovery~\cite{nakajima2013global}.

\section{Tensorizing Recurrent Neural Networks}
\label{section:rnn}
In this section, implementations of applying different tensor decomposition methods to RNN models are described. This section demonstrates the core steps of four tensor-decomposed RNN models. 1) Transform $W$ and $x$ into tensor representations, $\mathcal{W}$ and $\mathcal{X}$; 2) Decompose $\mathcal{W}$ into several low-rank tensors using different tensor decomposition methods; 3) the original product $W\cdot x$ is approximated by the tensor computation between decomposed weight tensor $\mathcal{W}$ and the input tensor $\mathcal{X}$.

\subsection{Tensorizing $W$, $x$ and $y$}
Considering $W$ is a 2-D matrix, it should be reshaped since tensor decomposition methods are mainly applied to high-order tensors. The input vector $x \in \mathbb{R}^N$, output vector $y \in \mathbb{R}^M$ and the weight matrix $W \in \mathbb{R}^{M \times N}$ should be tensorized into tensors $\mathcal{X} \in \mathbb{R}^{n_1 \times n_2 \times \cdots \times n_d}$, $\mathcal{Y} \in \mathbb{R}^{m_1 \times m_2 \times \cdots \times m_d}$ and $\mathcal{W} \in \mathbb{R}^{m_1 \times m_2 \times \cdots n_d \times n_1 \times n_2 \times \cdots \times n_d}$, respectively, where $M = \prod_{i=1}^d m_i$ and $N = \prod_{j=1}^d n_j$.

\subsection{Decomposing $W$}
In general, the key idea of building tensor-decomposed RNN is to transform the input-to-hidden weight matrices in different tensor decomposition forms.

\noindent \textbf{Tensor Train Decomposition} \quad
Following the Tensor Train Decomposition method, $\mathcal{W}$ can be decomposed into~\cite{yang2017tensor}:
\begin{equation}
\centering
TTD(\mathcal{W}) = \mathcal{G}_1 \mathcal{G}_2 \cdots \mathcal{G}_d.\nonumber
\end{equation}

Now instead of storing the full tensor $\mathcal{W}$ of size $N\cdot M$, the set of low-rank core tensors $\{ \mathcal{G} \}_{k=1}^d$ of size $\sum_{k=1}^d n_k \cdot m_k \cdot r_{k-1} \cdot r_k$ are stored.

\noindent \textbf{Tensor Ring Decomposition} \quad
Following the Tensor Ring Decomposition method, $\mathcal{W}$ can be decomposed into~\cite{pan2019compressing}:

\begin{equation}
\centering
\begin{split}
TRD(\mathcal{W}) = \sum_{r_0, \cdots, r_{2d-1}} &\mathcal{G}_{r_0, n_1, r_1}^{(1)} \cdots \mathcal{G}_{r_{d-1}, n_d, r_d}^{(d)} \cdot\\ &\mathcal{G}_{r_d, m_0, r_{d+1}}^{(d+1)} \cdots \mathcal{G}_{r_{2d-1}, m_d, r_0}^{(2d)} \nonumber
\end{split}
\end{equation}

For a $d$-order input and $d$-order output, the weight tensor is decomposed into $2d$ core tensors multiplied one by one where each corresponding to an input or an output dimension.

\noindent \textbf{Block Term Decomposition} \quad
Following the Block Term Decomposition method, $\mathcal{W}$ can be decomposed into~\cite{ye2018learning}:

\begin{equation}
\centering
BTD(\mathcal{W}) = \sum_{n=1}^N \mathcal{G}_n \bullet_1 \mathcal{A}_n^{(1)} \bullet_2 \cdots \bullet_d \mathcal{A}_d^{(d)}, \nonumber
\end{equation}
where $\mathcal{G} \in \mathbb{R}^{R_1 \times R_2 \times \cdots \times R_d}$ is the core tensor, $\mathcal{A}_n^{(d)} \in \mathbb{R} ^{N_d \times M_d \times R_d}$ is the factor tensor, $N$ denotes the CP-rank and $d$ denotes the Core-order.

\noindent \textbf{Hierarchical Tucker Decomposition} \quad
Following the Hierarchical Tucker Decomposition method, $\mathcal{W}$ can be decomposed into~\cite{yin2020compressing}:
\begin{equation}
\centering
\begin{split}
HTD(\mathcal{W}) = &\sum_{k=1}^{r_D} \sum_{p=1}^{r_{D_1}} \sum_{q=1}^{r_{D_2}} (\mathcal{G}_D)_{(k,p,q)} \\
&\cdot (\mathcal{U}_{D_1})_{(p, \phi_{D_1}(i, j))} (\mathcal{U}_{D_2})_{(q, \phi_{D_2}(i,j))}, \nonumber
\end{split}
\end{equation}
where the indices $i=(i_1, i_2, \cdots, i_d)$ and $j=(j_1, j_2, \cdots, j_d)$ are produced by the mapping function $\phi_s(i,j)$ for frame $\mathcal{U}_s$ with the given $s$ and $d$. Furthermore, $\mathcal{U}_{D_1}$ and $\mathcal{U}_{D_2}$ can be recursively calculated by:

\begin{equation}
\centering
\begin{split}
(\mathcal{U}_s)_{(k,\phi_s(i,j))} = &\sum_{p=1}^{r_{s_1}} \sum_{q=1}^{r_{s_2}} (\mathcal{G}_s)_{(k,p,q)} \cdot \\ &(\mathcal{U}_{s_1})_{(p,\phi_{s_1}(i,j))} (\mathcal{U}_{s_2})_{(q,\phi_{s_1}(i,j))}, \nonumber
\end{split}
\end{equation}
where $D=\{1, 2, \cdots, d\}$, $D_1 = \{1, \cdots, \lfloor d/2 \rfloor$ and $D_2 = \{\lceil d/2 \rceil, \cdots, d \}$ are associated with the left and right child nodes of the root node.

\subsection{Tensor Decomposition Layer}
After reshaping the weight matrix $W$ and the input vector $x$ into higher-order tensors and decomposing the weight tensor into tensor decomposed representations using the four different Tensor Decomposition methods as described above, the output tensor $\mathcal{Y}$ can be computed by manipulating $\mathcal{W}$ and $\mathcal{X}$. The final output vector $y$ can be obtained by reshaping the output tensor. The whole calculation from the input vector to the output vector can be denoted as the tensor decomposition layer (TDL):

\begin{equation}
y = TDL(W,x), \nonumber
\end{equation}
where TDL stands for one of Tensor Train Layer~\cite{yang2017tensor}, Tensor Ring Layer~\cite{pan2019compressing}, Block-Term Layer~\cite{ye2018learning} and HT Layer~\cite{yin2020compressing} depending on the tensor decomposition method used.

Tensor-decomposed RNN models can be obtained by replacing the multiplication between the weight matrix $W_h$ and input vector $x$ with TDL. The hidden state $h_t$ at time $t$ can be computed as:

\begin{equation}
\centering
h_t = \sigma (TDL(W_h, x_t) + U_h h_{t-1} + b), \nonumber
\end{equation}
where $\sigma(\cdot)$, $W_h$ and $U_h$ denote the activation function, the input-to-hidden layer weight matrix and the hidden-to-hidden layer matrix, respectively.

Replacing the multiplications between weight matrices and the input vector with TDL in the standard LSTM leads to:

\begin{equation}
\centering
\begin{split}
f_t &= \sigma(TDL(W_f, x_t) + U_f h_{t-1} + b_f)   \\
i_t &= \sigma(TDL(W_i, x_t) + U_i h_{t-1} + b_i)   \\
o_t &= \sigma(TDL(W_o, x_t) + U_o h_{t-1} + b_o)   \\
g_t &= tanh(TDL(W_g, x_t) + U_g h_{t-1} + b_g)   \\
c_t &= f_t \circ c_{t-1} + i_t \circ g_t   \\
h_t &= o_t \circ tanh(c_t), \nonumber
\end{split}
\end{equation}
where $\circ$, $\sigma(\cdot)$ and $tanh(\cdot)$ denote the Hadamard product (element-wise product), sigmoid function and the hyperbolic tangent function, respectively. The parameters $f_t$, $i_t$, $o_t$, $g_t$, $c_t$ and $h_t$ denote the forget gate's activation vector, input gate's activation vector, output gate's activation vector, cell input activation vector, cell state vector and hidden state vector, respectively. Note that $W_*$, $U_*$ and $b_*$ (where $*$ can be $f$, $i$, $o$ or $g$) are weight matrices and bias vector parameters which need to be learned during training. Back Propagation Through Time (BPTT) is used to compute the gradient of RNN~\cite{werbos1990backpropagation}. Following the regular RNN backpropagation processing, the gradient $\frac{\partial L}{\partial y}$ is computed by the original BPTT algorithm, where $y = W x_t$. Using the corresponding tensorization operation same to $y$, we can get the tensorized gradient $\frac{\partial L}{\partial \mathcal{Y}}$. All the tensorized RNN/LSTM models can be trained end-to-end directly. More details can be found in~\cite{yang2017tensor,pan2019compressing,ye2018learning,yin2020compressing}.

\section{Tensorizing Transformers}
\label{section:transformer}

\begin{figure}[htbp]
\centering
\resizebox{0.35\textwidth}{!}{%
\includegraphics{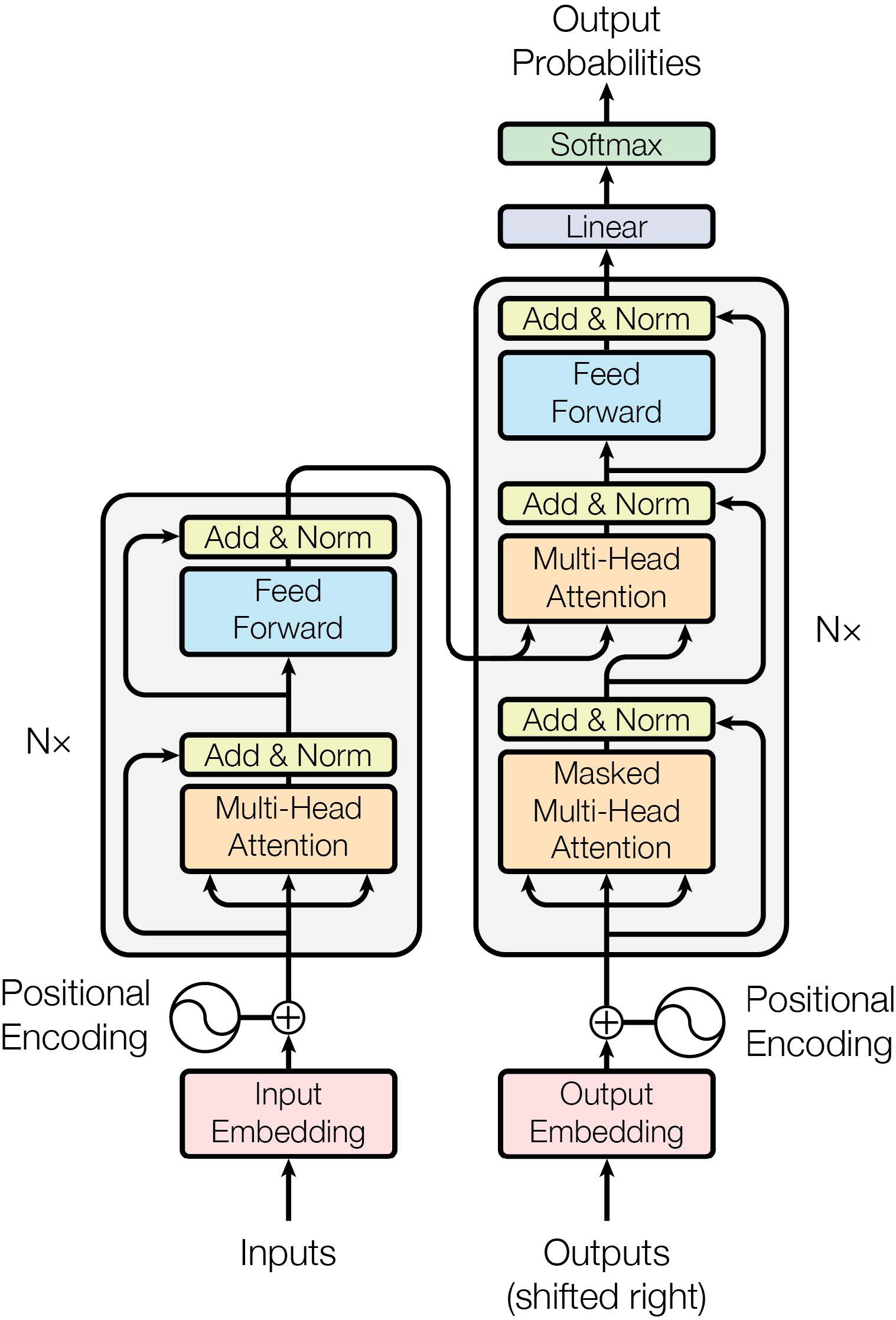}}
\caption{Model architecture of the Transformer~\cite{vaswani2017attention}.}
\label{fig:trans}
\end{figure}

In 2017, a team at Google Brain introduced the Transformer~\cite{vaswani2017attention} that gradually has become the preferred model for solving NLP problems, replacing RNN models such as LSTM~\cite{wolf2020transformers}. In this section, we review two tensor decomposition methods applied on Transformers: tensorized embedding layers~\cite{hrinchuk2019tensorized} and multi-linear attention~\cite{ma2019tensorized} that compress the embedding layers and the multi-head attention in Transformer, respectively.

\subsection{Multi-Linear Attention}

As shown in Fig.~\ref{fig:trans}, the original Transformer model utilized an encoder-decoder architecture. The encoder is made up of encoding layers that process the input successively, while the decoder consists of decoding layers that perform the same as the output of the encoder. Each encoder layer's function is to create encodings that contain information about which parts of the input are mutually relevant. Those encodings are passed to the next layer of the encoder as inputs. The reverse is done by each decoder layer, which takes all encodings and generates an output sequence utilizing the contextual information. Both encoder and decoder layers make use of an attention mechanism.

The attention function is called "Scaled Dot-Product Attention". It computes an output as a weighted sum of the values, with each value's weight determined by the compatibility function of the query with the corresponding key. In practice, the attention function is computed on a set of queries simultaneously by packing queries, keys and values into matrices $Q$, $K$ and $V$, respectively. Then the matrix of outputs can be computed as~\cite{vaswani2017attention}:
\begin{equation}
\centering
Attention(Q, K, V) = softmax(\frac{QK^T}{\sqrt{d}})V, \nonumber
\end{equation}
where $d$ is the number of columns of $Q$ and $K$. Instead of performing a single attention function, it is beneficial to linearly project the queries, keys and values $h$ times using distinct linear projections. The attention function is then applied concurrently to each of these projected versions of queries, keys, and values. The final values of the attention function with different inputs are then concatenated and projected to generate the final values. Multi-head attention allows the model to learn from different representations at different positions. It can be expressed as follows~\cite{vaswani2017attention}:
\begin{equation}
\centering
\begin{split}
MultiHead(Q, K, V) &= Concat(head_1, \cdots, head_h)W^O \\
\text{where } head_i &= Attention(QW^Q_i, KW^K_i, VW^V_i), \nonumber
\end{split}
\end{equation}
where matrices $W^Q_i$, $W^K_i$ and $W^V_i$ have the same size of $d_{model} \times d$ and $W^O \in \mathbb{R}^{hd \times d_{model}}$.

\begin{figure*}[htbp]
\centering
\resizebox{0.7\textwidth}{!}{%
\includegraphics{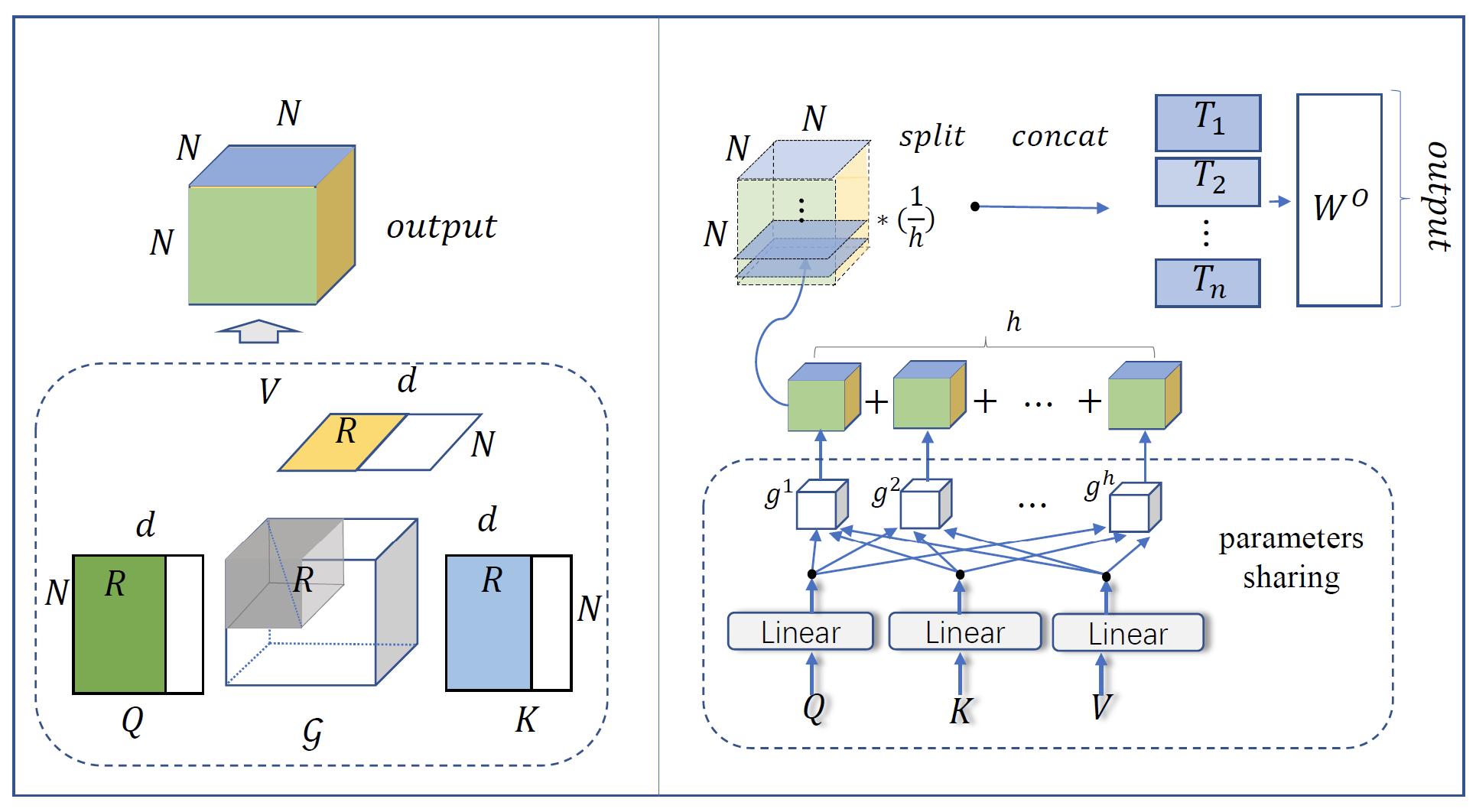}}
\caption{(left) Single-block attention using Tucker decomposition. (right) Multi-linear attention based on Block-Term tensor decomposition~\cite{ma2019tensorized}.}
\label{fig:trans1}
\end{figure*}

Fig.~\ref{fig:trans1}(left) shows the schematic diagram of Single-block attention based on the Tucker decomposition. The query, key and value can be mapped into three-factor matrices ($Q$, $K$ and $V$) where each factor matrix is composed of three groups of orthogonal basis vectors. Then a new attention can be constructed by initializing a 3rd-order diagonal tensor $\mathcal{G}$ as shown in Fig.~\ref{fig:trans1}(left) where $R$, $N$ and $d$ represent the rank of the tensor, the length of the input sequence and the dimension of the matrix, respectively. The Tucker-decomposition inspired Single-block attention can be computed as~\cite{ma2019tensorized}:
\begin{equation}
\centering
Atten_{TD}(\mathcal{G}; Q, K, V) = \sum_{i=1}^I \sum_{j=1}^J \sum_{m=1}^M \mathcal{G}_{ijm} Q_i \circ K_j \circ V_m, \nonumber
\end{equation}
where $\mathcal{G}$ is the trainable core tensor and $\circ$ represents the outer product. $Q_i$, $K_j$ and $V_k$ represent the column vectors of matrices $Q$, $K$ and $V$, respectively. Parameters $i$, $j$ and $m$ are the indexes of the core tensor. In practice, the core tensor $\mathcal{G}$ can be defined as~\cite{ma2019tensorized}:
\begin{equation}
\centering
\mathcal{G}_{ijk} = \left\{
\begin{aligned}
&rand(0, 1)  & i = j = m \\
&0 & otherwise \nonumber
\end{aligned}
\right.
\end{equation}
where $rand(0, 1)$ represent a random function. Then the Scaled Dot-Product attention can be computed by summing over the output of the Single-block attention function according to the second index as follows\cite{ma2019tensorized}:
\begin{equation}
\centering
Attention(Q, K, V)_{i, m} = \sum_{j = 1}^N Atten_{TD}(\mathcal{G}; Q, K, V)_{i, j, m}, \nonumber
\end{equation}
where $i$, $j$ and $m$ represent the indices of the output of the single-block attention.

In order to compress the multi-head function, multi-linear attention was proposed utilizing the idea of parameter sharing. As shown in Fig.~\ref{fig:trans1}(right), a set of linear projections map queries, keys and values to three matrices that can be composed of basis vectors. Then the multi-linear attention can be computed as follows~\cite{ma2019tensorized}:
\begin{equation}
    \centering
    \resizebox{.48\textwidth}{!}{$
    \begin{aligned}
        MultiLinear(\mathcal{G};Q,K,V) &= SplitConcat(\frac{1}{h} * (T_1 + \cdots + T_h))W^O \\
        \text{where } T_j &= Atten_{TD}(\mathcal{G}_j; QW^q, KW^k, VW^v), \nonumber
    \end{aligned}
    $}
\end{equation}
where the core tensor $\mathcal{G}_j$ is a diagonal tensor where $j \in \{1, \cdots, h\}$. $SplitConcat(\cdot)$ performs the concatenation following splitting. Matrices $W^q$, $W^k$ and $W^v$ are the parameter matrices that are shared.

\subsection{Tensorized Embedding Layers}
As shown in Fig.~\ref{fig:trans}, the embedding layers convert input words into vectors in NLP tasks. However, large vocabulary results in enormous weight matrices, which prevents their use in situations with constrained resources. A parameter-efficient embedding layer, TT-embedding, was introduced~\cite{hrinchuk2019tensorized}. The TT-embedding utilizes the Tensor Train decomposition by decomposing the huge embedding matrix into a sequence of much smaller 2-dimensional and 3-dimensional tensors. The method is the same as applying the Tensor Train decomposition to the fully connected layers of CNNs as described before.

Let $X \in \mathbb{R}$ be the embedding matrix of size $I \times J$ for a vocabulary of size $I$ and embedding dimension $J$. Given factorizations of its dimensions $I = \prod_{k=1}^N I_k$ and $J = \prod_{k=1}^N J_k$, the embedding matrix can be reshaped into a higher order tensor $\mathcal{X} \in \mathbb{R}^{I_1J_1 \times I_2J_2\times \cdots \times I_NJ_N}$. Then the high order tensor can be decomposed into a sequence of smaller tensors $\{\mathcal{G}_k \in \mathbb{R}^{R_{k-1}\times I_k \times J_k \times R_k} \}_{k = 1}^N$. The sequence $\{ R_k\}_{k=1}^{N-1}$ represent the TT-ranks that directly affect the compression ratio.

\section{Results}
\label{section:model}
\subsection{Convolutional Neural networks}
In this section, the implementations and results of applying these tensor decomposition methods to some classical CNN models are described. They are categorized based on the tensor decomposition methods.

\subsubsection{Tensor Train Decomposition}
In~\cite{garipov2016ultimate}, Tensor Train decomposition was applied to the convolutional layers and fully connected layers as described before. The proposed method was tested on CIFAR-10 dataset from~\cite{krizhevsky2009learning}. Two networks were built: one is dominated by the convolutional layers which occupy $99.54\%$ parameters of the network and the other is dominated by the fully connected layers which occupy $95.98\%$ parameters of the network. The first network consists of: conv with $64$ output channels; batch normalization; ReLU; conv with $64$ output channels; batch normalization; ReLU; max-pooling of size $3 \times 3$ with stride $2$; conv with $128$ output channels; batch normalization; ReLU; max-pooling of size $3 \times 3$ with stride $2$; conv with $128$ output channels; batch normalization; ReLU; conv with $128$ output channels; avg-pooling of size $4 \times 4$; FC of size ($128 \times 10$), where 'conv' and 'FC' stand for convolutional layer and fully connected layer, respectively. All convolutional filters were of size $3 \times 3$. Each convolutional layer excluding the first one was replaced by the TT-conv layer as mentioned before. The authors also compared the proposed Tensor Train decomposition method for the convolutional layer with the naive approach which directly applies the Tensor Train decomposition to the $4^{th}$-order convolutional kernel. As illustrated in TABLE~\ref{table:TT-1}, the proposed approach can achieve similar accuracies as the naive baseline on the $2 \times$ compression level. The second network was adapted from the first one by replacing the average pooling with two fully connected layers of size $8192 \times 1536$ and $1536 \times 512$. Initially, the fully-connected part was the memory bottleneck. As shown in TABLE~\ref{table:TT-2}, a $21.01 \times$ network compression with $0.7 \%$ accuracy drop can be achieved by only replacing the fully connected layers with TT-FC layers as described before. Then the bottleneck moves to the convolutional part. At this point, by additionally replacing the convolutional layers with the TT-conv layers, an $82.87 \times$ network compression with $1.1 \%$ accuracy drop can be achieved as shown in TABLE~\ref{table:TT-2}. More details can be found in~\cite{garipov2016ultimate}.

\begin{table}[htbp]
\centering
\caption{Compressing convolution-dominated CNN on the CIFAR-10 dataset~\cite{garipov2016ultimate}. Different rows with the same model name represent different choices of the TT-ranks. CR stands for compression rate.}
\begin{tabular}{ccc}
\hline
\textbf{Model}           & \textbf{Top-1 accuracy} & \textbf{CR}   \\ \hline
conv (original) & $90.7\%$           & -    \\ 
TT-conv         & $89.9\%$           & $2.02\times$ \\ 
TT-conv         & $89.2\%$           & $2.53\times$ \\ 
TT-conv         & $89.3\%$           & $3.23\times$ \\ 
TT-conv         & $88.7\%$           & $4.02\times$ \\ 
TT-conv (naive) & $88.3\%$           & $2.02\times$ \\ 
TT-conv (naive) & $87.6\%$           & $2.90\times$ \\ \hline
\end{tabular}
\label{table:TT-1}
\end{table}

\begin{table}[htbp]
\centering
\caption{Compressing FC-dominated CNN on the CIFAR-10 dataset~\cite{garipov2016ultimate}. Different rows with the same model name represent different choices of the TT-ranks. CR stands for compression rate.}
\begin{tabular}{ccc}
\hline
\textbf{Model}           & \textbf{Top-1 accuracy} & \textbf{CR}   \\ \hline
conv-FC (original) & $90.5\%$           & -    \\ \hline
conv-TT-FC         & $90.3\%$           & $10.72\times$ \\ 
conv-TT-FC         & $89.8\%$           & $19.38\times$ \\ 
conv-TT-FC         & $89.8\%$           & $21.01\times$ \\ 
TT-conv-TT-FC      & $90.1\%$           & $9.69\times$ \\ 
TT-conv-TT-FC      & $89.7\%$           & $41.65\times$ \\ 
TT-conv-TT-FC      & $89.4\%$           & $82.87\times$ \\ \hline
\end{tabular}
\label{table:TT-2}
\end{table}

\subsubsection{CP Decomposition}
In~\cite{lebedev2014speeding}, CP decomposition was implemented in two steps: applying CP decomposition to the convolutional layer using the NLS algorithm and then fine-tuning the entire network using backpropagation. Two network architectures were tested: small character-classification CNN from~\cite{jaderberg2014deep} and \texttt{AlexNet}~\cite{krizhevsky2012imagenet}. The character-classification CNN, called \texttt{CharNet}, has four convolutional layers. This CNN was used to classify images of size $24 \times 24$ into one of 36 classes (10 digits + 26 characters). Only the second and the third convolutional layers were compressed since they cost more than $90 \%$ of processing time. The second layer has $48$ input channels and $128$ output channels with filters of size $9 \times 9$. The third layer has $48$ input channels and $128$ output channels with filters of size $8 \times 8$. First, the second layer was compressed using CP decomposition with rank $64$. Then all layers but the new ones were fine-tuned to reduce the accuracy drop. Finally, the third layer was approximated using CP decomposition with rank $64$. Since the last approximation does not lead to a large accuracy drop, there is no need to fine-tune the network after that. The compressed network is $8.5$ times faster than the original one while the classification accuracy only drops by $1 \%$ to $90.2 \%$. \texttt{AlexNet} is one of the common object recognition networks and it has eight layers consisting of five convolution layers and three fully connected layers. The second convolutional layer of \texttt{AlexNet} was compressed in~\cite{lebedev2014speeding} by using CP decomposition. The running time of the second layer can be accelerated by $3.6 \times$ using a rank of $200$ at the expense of $0.5 \%$ accuracy degradation or by $4.5 \times$ with a rank of $140$ at the expense of $\approx 1 \%$ accuracy degradation. Another fact mentioned in~\cite{lebedev2014speeding} is that greedy CP decompositions like ALS work worse than NLS for CNN model compression.

\begin{figure*}[htbp]
\centering
\resizebox{0.8\textwidth}{!}{%
\includegraphics{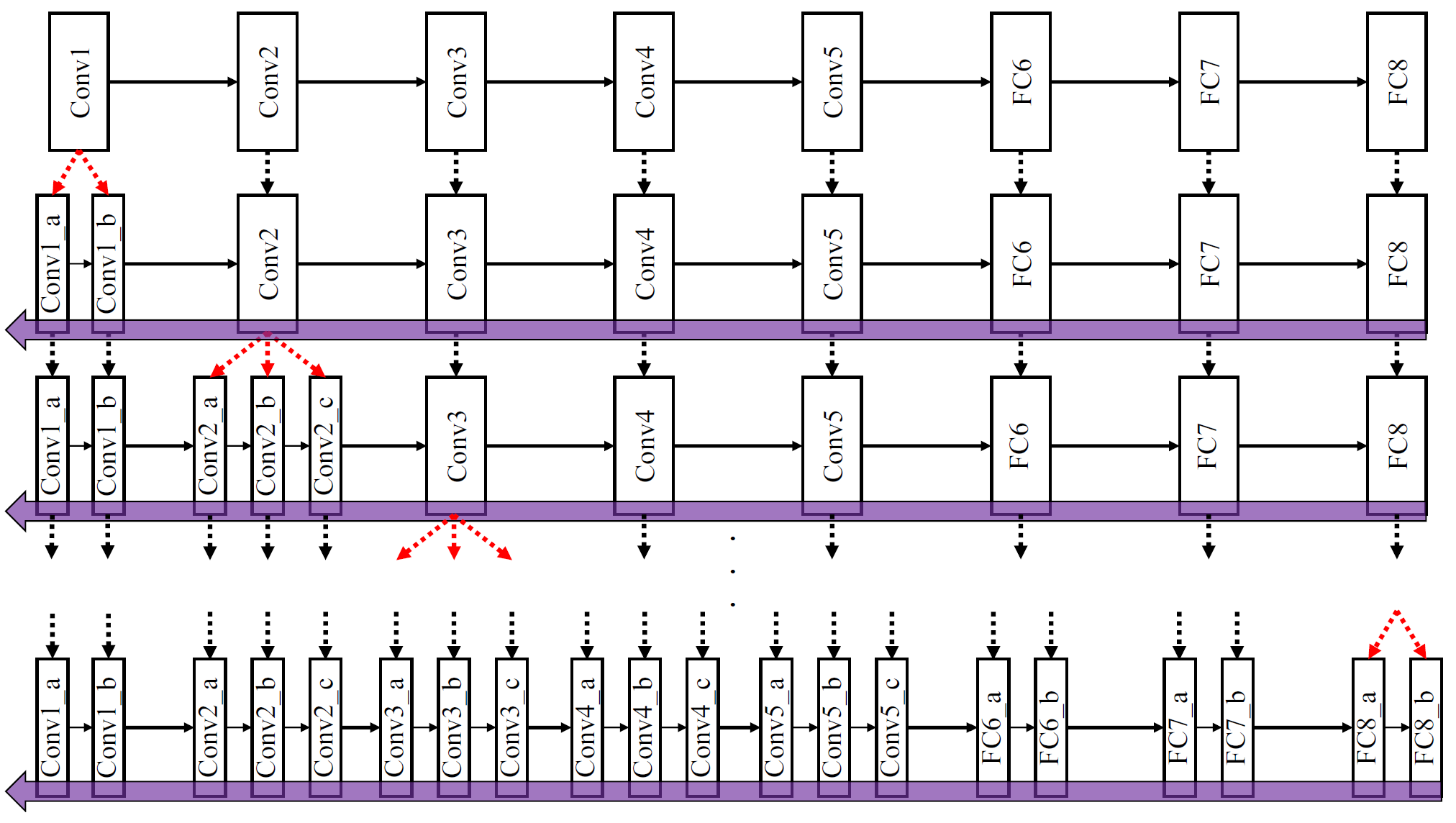}}
\caption{CP-TPM method in \texttt{AlexNet}~\cite{astrid2017cp}. The black solid arrow represents the connection between layers. The red dotted line stands for the decomposition process and the black dotted line means that the weights are taken from the previous iteration. The purple block arrow stands for fine-tuning by backpropagation to all the layers. First, Conv 1 is decomposed into 2 layers while the others remain the same. Then, fine-tuning is performed on the whole network. Afterward, Conv2 is decomposed into 3 layers and then fine-tuning is performed on the whole network followed. The process repeats until all the layers are decomposed and fine-tuned.}
\label{fig:tpm}
\end{figure*}

Tensor Power Method (TPM) \cite{allen2012sparse} was used to apply CP decomposition to the convolutional kernels in~\cite{astrid2017cp}. Compared to ALS, TPM can achieve the same variance with less rank since the rank-$1$ tensors found in the early steps of TPM represents most of the variances in the original tensor. TPM compressed the convolutional kernels by adding rank-1 tensors until a predefined number of rank-1 tensors is found. First, TPM finds a rank-$1$ tensor $\mathcal{K}_1$ by minimizing $||\mathcal{K} - \mathcal{K}_1||_2$. The next iteration approximates the residual tensor $\mathcal{K}_{residual} = \mathcal{K} - \mathcal{K}_1$ by minimizing $||\mathcal{K}_{residual} - \mathcal{K}_2||_2$. This continues until $\mathcal{K}_R$ is found. More details can be found in~\cite{allen2012sparse}. It is the first time that CP-based decomposition is applied to the whole convolutional layer in~\cite{astrid2017cp}. \texttt{AlexNet} was used here. The authors overcome the instability of CP decomposition by fine-tuning after each layer's decomposition. The fully connected layers were decomposed using SVD as described before. Fig.~\ref{fig:tpm} from~\cite{astrid2017cp} shows that decomposition and fine-tuning are performed iteratively from Conv1 to FC8. Black solid arrows represent the connection between each layer. Red dotted lines represent the decomposition processes. Black dotted lines show that the weights do not change from the previous iteration. Purple block arrows represent fine-tuning by backpropagation to the whole network. The rank of each layer was set to be proportional to its sensitivity which is defined as \textit{loss/total\_loss}. This method achieved $6.98 \times$ parameter reduction and $3.53 \times$ running time reduction with the expense of $1.42 \%$ accuracy loss.

\subsubsection{Tucker Decomposition}

\begin{figure}[htbp]
\centering
\resizebox{0.485\textwidth}{!}{%
\includegraphics{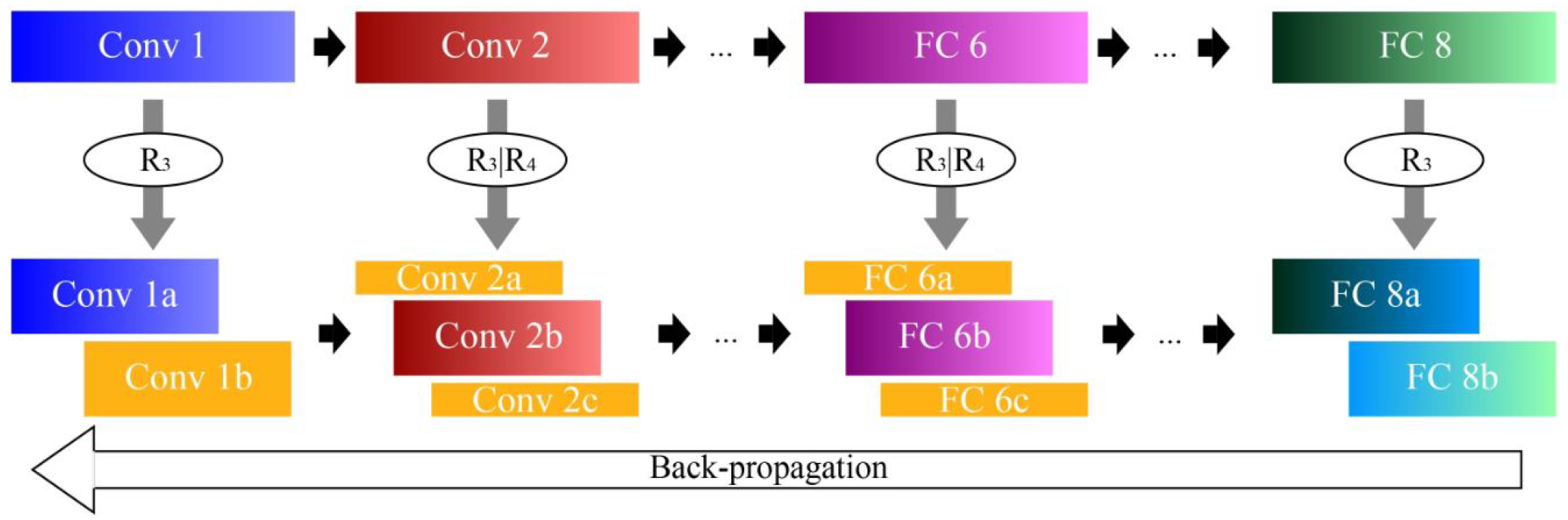}}
\caption{One-shot whole network compression based on Tucker decomposition~\cite{kim2015compression}. Tucker-2 decomposition is applied from the second convolutional layer to the first fully connected layers while Tucker-1 decomposition is applied to the other layers.}
\label{fig:tucker}
\end{figure}

In~\cite{kim2015compression}, one-shot Tucker Decomposition on the whole network consists of three steps: rank selection using VBMF, Tucker decomposition on each layer's kernel tensor and one-shot fine-tuning the whole network with standard back-propagation. Fig.~\ref{fig:tucker} from~\cite{kim2015compression} shows the whole scheme. The accuracy significantly dropped after step two but recovered quickly in one epoch. Four representative CNNs, \texttt{AlexNet}, \texttt{VGG-S}, \texttt{GoogLeNet} and \texttt{VGG-16} were compressed using Tucker decomposition in~\cite{kim2015compression}. For \texttt{GoogLeNet}, only the $3 \times 3$ convolution kernel was compressed in the case of \texttt{inception} module. For \texttt{VGG-16}, only the convolutional layers were compressed. This method achieved $5.46 \times$ / $2.67 \times$ (\texttt{AlexNet}), $7.40 \times$ / $4.80 \times$ (\texttt{VGG-S}), $1.28 \times$ / $2.06 \times$ (\texttt{GoogLeNet}) and $1.09 \times$ / $4.93 \times$ (\texttt{VGG-16}) reductions in total weights and FLOPs, respectively.

\subsection{Recurrent Neural Networks}
In this section, the results of applying different tensor decomposition methods to RNN models are described.

\begin{table}[htbp]
\centering
\caption{Comparison of time complexity and space complexity of vanilla RNN, Tensor Train decomposed RNN (TT-RNN)~\cite{yang2017tensor}, Tensor Ring decomposed RNN (TR-RNN)~\cite{pan2019compressing}, Block-Term decomposed RNN (BT-RNN)~\cite{ye2018learning} and Hierarchical Tucker decomposed RNN (HT-RNN)~\cite{yin2020compressing}. The weight matrix's shape is $M \times N$. The input and hidden tensors' shapes are $N = n_1 \times \cdots \times n_d$ and $M = m_1 \times \cdots \times m_d$, respectively. Here, $n = max(n_k)$ for $k \in [1, d]$ and $m = max(m_k)$ for $k \in [1, d]$. All tensor decomposed models are set in the same rank $R$. $C$ is the CP-rank defined in BT-RNN model.}
\begin{tabular}{lll}
\hline
\textbf{Method}          & \textbf{Time} & \textbf{Space} \\ \hline
RNN forward     & $\mathcal{O}(NM)$   &   $\mathcal{O}(NM)$    \\
RNN backward    & $\mathcal{O}(NM)$   &   $\mathcal{O}(NM)$    \\ \hline
TT-RNN forward  & $\mathcal{O}(dmR^2N)$   &  $\mathcal{O}(RM)$     \\
TT-RNN backward & $\mathcal{O}(d^2mR^4N)$ &  $\mathcal{O}(R^3M)$     \\ \hline
TR-RNN forward  & $\mathcal{O}(dR^3N+dR^3M)$        &   $\mathcal{O}(dmnR^2)$    \\
TR-RNN backward & $\mathcal{O}(d^2R^5N+nd^2R^5M)$   &   $\mathcal{O}(dmnR^2)$    \\ \hline
BT-RNN forward  & $\mathcal{O}(dmR^dNC)$     &   $\mathcal{O}(R^dM)$    \\
BT-RNN backward & $\mathcal{O}(d^2mR^dNC)$   &   $\mathcal{O}(R^dM)$    \\ \hline
HT-RNN forward  & $\mathcal{O}(dmR^2N+dR^3N)$     &   $\mathcal{O}(dmnR+dR^3)$    \\
HT-RNN backward & $\mathcal{O}(d^2mr^5N+d^2R^6N)$ &   $\mathcal{O}(dmnR+dR^3)$    \\ \hline
\end{tabular}
\label{table:comp}
\end{table}

To better understand the impact of different tensor decomposition methods on RNNs, the time complexity and space complexity of different RNN models are summarized in Table~\ref{table:comp}. Notice that HT-RNN has the lowest space complexity and TT-RNN has the lowest time complexity.

The UCF11 YouTube Action dataset~\cite{liu2009recognizing} contains $1600$ video clips of a resolution $320 \times 240$, falling into $11$ action categories (e.g., basketball, biking, diving, etc.). Each category contains $25$ video groups, within each contains more than $4$ clips. Table~\ref{table:ucf11} summarizes the performance of different tensor-decomposed LSTM models on UCF11 dataset. Notice that HT-LSTM requires at least $1.38 \times$ fewer parameters with at least $0.3 \%$ increase in accuracy compared to the other tensor-decomposed LSTM models.

\begin{table}[htbp]
\caption{Comparison of different tensor-decomposed LSTM models on UCF11 dataset. CR stands for compression ratios.}
\centering
\begin{tabular}{cccc}
\hline
\textbf{Method}  & \textbf{CR}    & \textbf{\# of Param.} & \textbf{Accuracy} \\ \hline
LSTM    & $1$     & 59M          & 69.7 \%  \\
TT-LSTM~\cite{yang2017tensor} & $17554 \times$ & 3360         & 79.6 \%  \\ 
TR-LSTM~\cite{pan2019compressing} & $34193 \times$ & 1725         & 86.9 \%  \\
BT-LSTM~\cite{ye2018learning} & $17414 \times$ & 3387         & 85.3 \%  \\ 
HT-LSTM~\cite{yin2020compressing} & $47375 \times$ & 1245         & 87.3 \%  \\ \hline
\end{tabular}
\label{table:ucf11}
\end{table}

\subsection{Transformers}
In this section, the results of applying different tensor decomposition methods to Transformers are described.

\subsubsection{Multi-Linear Attention}
The proposed attention method was tested on three language modeling tasks (PTB, WikiText-103 and One-billion) and a neural machine translation task (WMT-2016 English-German)~\cite{ma2019tensorized}.

Language modeling is the task of predicting the next word or character in a document. It can be used to generate text or further fine-tuned to solve different NLP tasks~\cite{radford2018improving}. Three datasets were chosen: one of small size (PTB), one of medium size (WikiText-103) and one of large size (One-billion). PTB contains 929,900 training tokens, 73,900 validation words, and 82,900 test words~\cite{deoras2011empirical}. WikiText-103 has 267,735 distinct tokens. The dataset is a long-term dependency word-level language modeling benchmark. It contains 103 million training tokens from 28 thousand articles, with an average length of 3.6 thousand tokens per article. The One-Billion Word benchmark is a large dataset with 829,250,940 tokens over a vocabulary of 793,471 words. Models were evaluated based on the average per-word log-probability, Perplexity (PPL). The lower the PPL, the more accurate the model. The standard multi-head attention layers in Transformer were replaced with Multi-linear attention. A comparison of different model configurations on different datasets is shown in Table~\ref{table:ob} and Table~\ref{table:pw}. Notice that the tensorized transformer with Multi-linear attention achieves lower PPL with much fewer parameters than other models in the three datasets.

\begin{table}[htbp]
\caption{Comparison of different model configurations on One-Billion~\cite{ma2019tensorized}. Tensorized Transformer is the model with Multi-linear attention. Core-1 denotes a model with a single-block term tensor, whereas core-2 denotes a model with two block term tensors.}
\centering
\begin{tabular}{ccc}
\hline
\textbf{Model}                & \textbf{Params} & \textbf{Test PPL} \\ \hline
RNN-1024+9 Gram~\cite{chelba2013one}              & 20B             & 51.3              \\
LSTM-2018-512~\cite{jozefowicz2016exploring}                 & 0.83B           & 43.7              \\
GCNN-14 bottleneck~\cite{dauphin2017language}            & -               & 31.9              \\
LSTM-8192-1024+CNN Input~\cite{jozefowicz2016exploring}      & 1.04B           & 30.0              \\ \hline
High-Budget MoE~\cite{shazeer2017outrageously}               & 5B              & 28.0              \\
LSTM+Mos~\cite{yang2017breaking}                      & 113M            & 37.10             \\
Transformer+adaptive input~\cite{baevski2018adaptive}    & 0.46B           & 23.7              \\
Transformer-XL Base~\cite{dai2019transformer}          & 0.46B           & 23.5              \\
Transformer-XL Large~\cite{dai2019transformer}          & 0.8B            & 21.8              \\
Tensorized Transformer core-1~\cite{ma2019tensorized} & 0.16B           & 20.5              \\
Tensorized Transformer core-2~\cite{ma2019tensorized} & 0.16B           & 19.5              \\ \hline
\end{tabular}
\label{table:ob}
\end{table}

\begin{table*}[htbp]
\caption{Comparison of different model configurations on PTB and WikiText-103~\cite{ma2019tensorized}. Tensorized Transformer is the model with Multi-linear attention. Core-1 denotes a model with a single-block term tensor, whereas core-2 denotes a model with two block term tensors.}
\centering
\begin{tabular}{cccc|ccc}
\hline
\multirow{2}{*}{\textbf{Model}} & \multicolumn{3}{c|}{\textbf{PTB}} & \multicolumn{3}{c}{\textbf{WikiText-103}} \\ \cline{2-7} 
                                & Params   & Val PPL   & Test PPL   & Params      & Val PPL      & Test PPL     \\ \hline
LSTM+augmented loss~\cite{inan2016tying}             & 24M      & 75.7      & 48.7       & -           & -            & 48.7         \\
Variational RHN~\cite{zoph2016neural}                 & 23M      & 67.9      & 65.4       & -           & -            & 45.2         \\
4-layer QRNN~\cite{merity2016pointer}                    & -        & -         & -          & 151M        & -            & 33.0         \\
AWD-LSTM-MoS~\cite{yang2017breaking}                    & 22M      & 58.08     & 55.97      & -           & 29.0         & 29.2         \\ \hline
Transformer+adaptive input~\cite{baevski2018adaptive}      & 24M      & 59.1      & 57         & 247M        & 19.8         & 20.5         \\
Transformer-XL-Base~\cite{dai2019transformer}             & 24M      & 56.72     & 54.52      & 151M        & 23.1         & 24.0         \\
Transformer-XL-Large~\cite{dai2019transformer}            & -        & -         & -          & 257M        & -            & 18.3         \\
Transformer-XL+TT~\cite{hrinchuk2019tensorized}               & 18M      & 57.9      & 55.4       & 130M        & 23.61        & 25.70        \\
Sparse Transformer~\cite{child2019generating}              & 14M      & 74.0      & 73.1       & 174M        & 38.98        & 40.23        \\
Tensorized Transformer core-1~\cite{ma2019tensorized}   & 12M      & 60.5      & 57.9       & 85.3M       & 22.7         & 20.9         \\
Tensorized Transformer core-2~\cite{ma2019tensorized}   & 12M      & 54.25     & 49.8       & 85.3M       & 19.7         & 18.9         \\ \hline
\end{tabular}
\label{table:pw}
\end{table*}

Neural Machine Translation involves translating text or speech from one language to another. The baseline is a vanilla Transformer trained on WMT 2016 English-German dataset~\cite{sennrich2016edinburgh}. For comparison, each of the attention layers was replaced with Multi-linear attention in the Encoder of the Transformer. The results are summarized in Table~\ref{table:WMT}. Notice that tensorized transformers with Multi-linear attention achieve better performance with fewer parameters than the vanilla Transformer.

\begin{table}[htbp]
\caption{Comparison of different Transformers on WMT-16 English-to-German translation~\cite{ma2019tensorized}. Tensorized Transformer is the model with Multi-linear attention. Core-1 denotes a model with a single-block term tensor, whereas core-2 denotes a model with two block term tensors.}
\centering
\begin{tabular}{ccc}
\hline
\textbf{Model}                & \textbf{Params} & \textbf{BLEU} \\ \hline
Transformer~\cite{vaswani2017attention}                   & 52M             & 34.5          \\
Tensorized Transformer core-1 & 21M             & 34.10         \\
Tensorized Transformer core-2 & 21.2M           & 34.91         \\ \hline
\end{tabular}
\label{table:WMT}
\end{table}

\subsubsection{Tensorized Embedding Layers}
The proposed TT-embedding layer was tested on two language modeling tasks (PTB and WikiText-103) and a machine translation task (WMT 2014 English–German). As shown in Table~\ref{table:pw}, Transformer-XL+TT stands for the transformers with TT-embedding layers. Compared to the Transformer with Multi-linear attention, Transformer-XL+TT can not achieve that high compression ratio. For the machine translation task, the baseline is a Transformer-big model on WMT 2014 English-German dataset~\cite{vaswani2017attention}. This dataset has around 4.5 million sentence pairs. The results are summarized in Table~\ref{table:WMT14}. Notice that the embedding layers can be compressed significantly at the cost of a small drop in the BLEU scores.

\begin{table*}[htbp]
\caption{Comparison of different Transformers on WMT-14 English-to-German translation~\cite{hrinchuk2019tensorized}. Both case-sensitive tokenized BLEU (higher is better) and de-tokenized SacreBLEU~\cite{post2018call} are reported. Emb compr. stands for the compression rate for the embedding layers.}
\centering
\begin{tabular}{ccccccc}
\hline
\textbf{Model}               & \textbf{Embedding shape}         & \textbf{TT-Rank} & \textbf{Emb compr.} & \textbf{Total params} & \textbf{Token BLEU} & \textbf{Scare BLEU} \\ \hline
Transformer-Big~\cite{vaswani2017attention}     & $32768 \times 1024$            & -   &  1  & 210M   & 29.58      & 28.84      \\
Transformer-Big+TT1~\cite{hrinchuk2019tensorized} & $(32, 32, 32)\times (8, 8, 16)$ & 64 & 15.3     & 179M   & 29.17      & 28.53      \\
Transformer-Big+TT2~\cite{hrinchuk2019tensorized} & $(32, 32, 32)\times (8, 8, 16)$ & 48 & 26.8     & 178M   & 28.53      & 26.8       \\
Transformer-Big+TT2~\cite{hrinchuk2019tensorized} & $(32, 32, 32)\times (8, 8, 16)$ & 32 & 58.5     & 177M   & 28.26      & 27.70 \\ \hline     
\end{tabular}
\label{table:WMT14}
\end{table*}

% \section{Future Directions}
% \label{section:future}

\section{Conclusion}
\label{section:conclusion}
This paper has summarized different tensor decomposition methods that are used to compress model parameters of CNNs, RNNs and Transformers. In particular, three decomposition methods for CNNs, four decomposition methods for RNNs and two decomposition methods for Transformers have been reviewed. Finding the best rank for the low-rank approximation remains a challenge and has to be found by trial and error. The model compression ratio is dependent on the rank approximation and the performance degradation that can be tolerated. For RNNs, the model compression ratio is much higher than those of CNNs. Additionally, the accuracy of the compressed RNN model can be better than the original model. All these models are trained end-to-end using backpropagation instead of being obtained by applying tensor decomposition methods to the pre-trained standard models. It is possible for them to achieve higher accuracy than the standard models with fewer parameters.

Future work needs to be directed toward comparing the performance of various decomposition methods using common datasets. Hardware-aware Tucker Decomposition was proposed to efficiently generate highly accurate and compact CNN models on GPUs~\cite{xiang2022tdc}. Algorithm and Hardware co-Design of high-performance energy-efficient LSTM networks was introduced based on Hierarchical Tucker Tensor Decomposition~\cite{gong2022algorithm}. Future research also needs to be directed toward the evaluation of the hardware performance of these decomposition methods.
%While this paper has addressed inference, future work will need to be directed towards reducing time complexity of training the models that are compressed by tensor decomposition.

%

% \appendices
% \section{Proof of the First Zonklar Equation}
% Appendix one text goes here.

% % you can choose not to have a title for an appendix
% % if you want by leaving the argument blank
% \section{}
% Appendix two text goes here.

% % use section* for acknowledgment
% \section*{Acknowledgment}

% The authors would like to thank...

% Can use something like this to put references on a page
% by themselves when using endfloat and the captionsoff option.
\ifCLASSOPTIONcaptionsoff
  \newpage
\fi

% trigger a \newpage just before the given reference
% number - used to balance the columns on the last page
% adjust value as needed - may need to be readjusted if
% the document is modified later
%\IEEEtriggeratref{8}
% The "triggered" command can be changed if desired:
%\IEEEtriggercmd{\enlargethispage{-5in}}

% references section

% can use a bibliography generated by BibTeX as a .bbl file
% BibTeX documentation can be easily obtained at:
% http://mirror.ctan.org/biblio/bibtex/contrib/doc/
% The IEEEtran BibTeX style support page is at:
% http://www.michaelshell.org/tex/ieeetran/bibtex/
% \vspace{1em}

% \noindent \textbf{Acknowledgment}: This research was supported in part by the National Science Foundation under grant number CCF-1954749.

\bibliographystyle{IEEEtran}
\bibliography{references}

\begin{IEEEbiography}[{\includegraphics[width=1in,height=1.25in,clip,keepaspectratio]{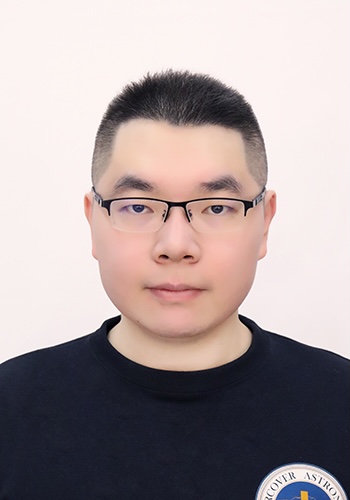}}]{Xingyi Liu}
(S’18) received the B.S. degree from the University of Minnesota-Twin Cities, Minneapolis, MN, USA, in 2016, the M.S.E.E degree from the University of Minnesota-Twin Cities, Minneapolis, MN, USA, in 2018. Currently, he is working towards the Ph.D. degree in the Department of Electrical and Computer Engineering, University of Minnesota-Twin Cities, Minneapolis, MN, USA. His research interests include molecular signal processing, DNA computing and deep neural network using stochastic logic.
\end{IEEEbiography}

\begin{IEEEbiography}[{\includegraphics[width=1.1in,keepaspectratio]{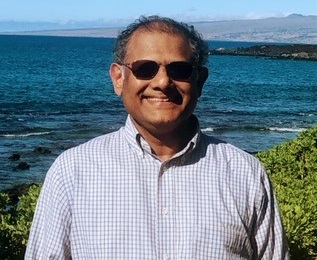}}]{Keshab K. Parhi}
(S'85-M'88-SM'91-F'96) received the B.Tech. degree from Indian Institute of Technolgy, Kharagpur, India, in 1982, the M.S.E.E. degree from the University of Pennsylvania, Philadelphia, PA, USA, in 1984, and the Ph.D. degree from the University of California at Berkeley, Berkeley, CA, USA, in 1988.

He has been with the University of Minnesota, Minneapolis, MN, USA, since 1988, where he is currently the Erwin A. Kelen Chair and a Distinguished McKnight University Professor in the Department of Electrical and Computer Engineering. He has published 700 papers, is the inventor or co-inventor of 34 patents, has authored the textbook \textit{VLSI Digital Signal Processing Systems} (New York, NY, USA: Wiley, 1999), and coedited the reference book \textit{Digital Signal Processing for Multimedia Systems} (Boca Raton, FL, USA: CRC Press, 1999). His current research interests include the VLSI architecture design of signal processing and machine learning systems, neuromorphic computing, data-driven neuroscience and biomarkers for neuro-psychiatric disorders, hardware security and DNA computing. 

Dr. Parhi has served as the Editor-in-Chief of the IEEE TRANSACTIONS ON CIRCUITS AND SYSTEMS PART I from 2004 to 2005.  He currently serves on the Editorial Board of the \textit{Journal of Signal Processing Systems} (Springer). He has served as the Technical Program Co-Chair of  the 1995 IEEE VLSI Signal Processing Workshop and the 1996 Application Specific Systems, Architectures, and Processors (ASAP) conference, and as the General Chair of the 2002 IEEE Workshop on Signal Processing Systems (SiPS). He was the Distinguished Lecturer of the IEEE Circuits and Systems Society during 1996-1998 and 2019-2021. He served as a Board of Governors Elected Member of the IEEE Circuits and Systems Society from 2005 to 2007. He is the recipient of numerous awards including the 2017 Mac Van Valkenburg award, the 2012 Charles A. Desoer Technical Achievement award and the 1999 Golden Jubilee medal, from the IEEE Circuits and Systems society, the 2013 Distinguished Alumnus Award from IIT Kharagpur, the 2013 Graduate/Professional Teaching Award from the University of Minnesota, the 2004 F. E. Terman award from the American Society of Engineering Education, the 2003 IEEE Kiyo Tomiyasu Technical Field Award, and the 2001 IEEE W. R. G. Baker Prize Paper Award. He is a Fellow of ACM, AAAS, AIMBE and the National Academy of Inventors.
\end{IEEEbiography}

\end{document}